\documentclass[preprint,12pt]{elsarticle}
\usepackage[margin=1in]{geometry}
\usepackage{graphicx}
\usepackage{overpic}
\usepackage{subcaption}
\usepackage{booktabs}
\usepackage{multirow}
\usepackage{adjustbox}
\usepackage{colortbl}
\usepackage{amsmath,amssymb,amsthm,amsfonts}
\usepackage{xcolor}
\usepackage{soul}
\usepackage{enumitem}
\usepackage{pifont}
\usepackage{lineno}
\usepackage{footnote}
\usepackage{tikz}
\usepackage[normalem]{ulem}
\usetikzlibrary{calc}
\usepackage{hyperref}
\usepackage{cleveref}
\biboptions{numbers,comma,round,square}

\journal{Elsevier}
\begin{document}

\title{Physics-Integrated Operator Learning via Gaussian Splatting Representations}
\author{Jihao Zhang}
\author{Junyi Guo}
\author{Jian-Xun Wang\corref{corxh}}
\ead{jw2837@cornell.edu}

\address{Sibley School of Mechanical and Aerospace Engineering, Cornell University, Ithaca, NY, USA}
\cortext[corxh]{Corresponding author. Tel: +1 540 3156512}

\begin{abstract}
Neural operators provide efficient surrogates for spatiotemporal PDE systems, but purely data-driven formulations often accumulate substantial errors during long-horizon autoregressive prediction and may fail to exploit available governing-equation structure. Existing approaches incorporate physics primarily through residual-based training objectives or PDE-specific architectural constraints, which can introduce optimization difficulties or limit architectural generality. In this work, we introduce a representation-level approach to physics integration in which a feed-forward Gaussian splatting (FFGS) representation serves as a continuous interface between discretized solution fields and governing operators. The FFGS representation reconstructs the state as a continuous Gaussian field with closed-form spatial derivatives, allowing available physical PDE operators to be integrated directly within the learned evolution map without introducing a physics-residual loss. We evaluate the framework across two- and three-dimensional PDE systems, including advection, diffusion, nonlinear self-advection, and reaction dynamics. Over long-horizon autoregressive rollouts, the proposed framework reduces relative $\ell_2$ error by $1.5\times$--$2.2\times$ compared with the strongest purely data-driven baseline across the benchmark suite, while consistently improving spectral fidelity. The framework also remains effective when the governing equations are partially known, demonstrating robustness to incomplete physics. These results demonstrate that continuous field representations can provide a practical interface for incorporating known physical structure into generic neural-operator surrogates.
\end{abstract}


\maketitle

\section{Introduction}
\label{sec:intro}

Spatiotemporal dynamics governed by partial differential equations (PDEs) are ubiquitous in science and engineering, from convective and diffusive transport to nonlinear wave steepening and moving reaction interfaces. Traditional numerical methods discretize the governing operator and advance the state by time integration; they resolve these dynamics accurately, but at a cost that becomes prohibitive in many-query settings such as design optimization, inverse problems, data assimilation, and uncertainty quantification, where the forward model must be evaluated repeatedly across different initial conditions and physical parameters. This has motivated a rapidly growing effort to build data-driven surrogates that learn a fast approximation of the solution operator from precomputed trajectories and, once trained, stand in for the numerical solver at a small fraction of its cost. 

Neural operators (NOs)~\cite{lu_deeponet_2021, kovachki_neural_2024}, which use neural networks to parameterize mappings between infinite-dimensional function spaces, are currently the leading class of such surrogates. They have demonstrated strong predictive performance for a broad range of spatiotemporal systems, including weather dynamics~\cite{pathak2022fourcastnet,bi2023accurate} and turbulent flows~\cite{li_fnoforles_2022,wang_fnochannelflow_2024}.  However, standard NOs are primarily trained from trajectory data and infer the underlying dynamics implicitly from finite observations. This limitation becomes particularly evident in autoregressive prediction, where local approximation errors accumulate through repeated operator evaluations and progressively degrade long-horizon accuracy~\cite{lippe_pderefiner_2023}. The degradation is further amplified when the deployed dynamics deviate from the training distribution~\cite{takamoto_pdebench_2024,benitez2024out}. These failure modes motivate the integration of governing-equation knowledge into operator learning.

A direct strategy is to augment supervised operator learning with auxiliary loss terms derived from the governing equations~\cite{wang_pideeponet_2021,li_physics-informed_2023}. Such physics-informed formulations inherit several optimization difficulties of PINNs~\cite{raissi_pinn_2019}: the data and physics objectives exhibit disparate scales and gradient dynamics, making their relative weighting delicate and problem-dependent~\cite{wang_pinnfail_2022,krishnapriyan_failure_2021}. Moreover, enforcing the governing equations through a residual requires repeated evaluation of spatial and temporal derivatives during training. For coordinate-based representations, these derivatives are commonly obtained through automatic differentiation, whose computational and memory costs increase with the number of collocation points and derivative order, thereby limiting the scalability of residual-based physics integration.

An alternative direction incorporates physical knowledge directly into the construction of the operator. For example, conservation-preserving neural operators enforce prescribed invariants through differential constructions~\cite{Liu2024clawNO}, while invariant and equivariant operators encode symmetry properties of the governing system~\cite{liu_ino_2023,helwig_gfno_2023}. Other approaches introduce local differential interactions or physics-motivated feature transformations into neural-operator layers~\cite{hu2025physics,liu_NOlocalizedint_2024}. These methods demonstrate that physical knowledge can improve learned dynamics when incorporated beyond the training objective. However, the corresponding constructions are often tailored to specific PDE systems, and incorporating a different governing equation may require redesigning the corresponding physics-aware components.

This PDE-specific design motivates introducing governing knowledge through the field representation rather than through the neural operator architecture. Such a representation must both reconstruct the physical field from its discretized state and provide direct access to the spatial differential quantities required by the governing operator. Explicit continuous basis representations are well suited to this role because both the reconstructed field and its spatial derivatives can be evaluated analytically from the basis functions. The available governing terms can then be evaluated directly on this continuous representation while retaining a generic neural operator architecture.

The use of explicit continuous basis representations is well established in classical radial basis function (RBF) collocation methods, where the solution is represented by smooth basis functions and differential operators are applied analytically to the basis to enforce the governing equations at collocation points~\cite{kansa_rbf_1990,franke_rbfcollocation_1998}. More recently, Gaussian primitives obtained by 3D Gaussian Splatting~\cite{kerbl_3dGS_2023} have been explored for physical-field representation and modeling. Existing studies span Gaussian representations for compact field encoding~\cite{shenoy2026gaussian}, Gaussian-basis operator learning~\cite{li2026basis}, and PDE-constrained Gaussian optimization~\cite{xing_gaussian_2025}.

The original Gaussian-splatting paradigm, however, relies on per-instance optimization of the primitive parameters. Applying such a procedure to every evolving PDE state would introduce substantial overhead in autoregressive time stepping. Feed-forward Gaussian-splatting (FFGS) methods developed in computer vision provide an efficient alternative by amortizing Gaussian construction into learned models that predict primitive parameters in a single forward pass~\cite{charatan_pixelsplat_2024,wang_volsplat_2025,zhang_sparsesplat_2026}. Related amortized Gaussian representations have also begun to appear in scientific applications, for example for reconstructing physical fields from sparse observations~\cite{huang2026fluidsplat}. However, the use of an amortized Gaussian representation as a continuous interface for evaluating governing operators within neural-operator time evolution remains largely unexplored.

To address this gap, we develop a physics-integrated neural-operator framework that employs FFGS as a continuous representation of physical fields. The analytic Gaussian representation provides direct access to the spatial derivatives required to evaluate available governing terms, allowing their contribution to be incorporated into the learned evolution map while retaining a generic neural-operator backbone. We further extend the framework to a partial-physics setting, where the governing form is known but its coefficients are identified from trajectory data. We evaluate the framework across two- and three-dimensional systems involving advection, diffusion, nonlinear self-advection, and reaction dynamics. Across these settings, representation-level physics integration consistently improves long-horizon rollout accuracy and spectral fidelity over purely data-driven baselines, with substantial gains retained when only partial governing information is available. Component ablations further show that the embedded-physics and neural-operator components play complementary roles in the resulting dynamics. Overall, these results establish FFGS as a practical continuous interface for incorporating fully or partially known governing structure into generic neural-operator surrogates.

The remainder of this paper is structured as follows. In Sec.~\ref{sec:method}, we introduce the physics-integrated neural-operator framework and formulate the FFGS representation, embedded-physics component, and rollout-training procedure. In Sec.~\ref{sec:setup}, we describe the benchmark problems, training protocol, baseline models, and evaluation metrics. In Sec.~\ref{sec:results}, we report the numerical results and evaluate long-horizon rollout accuracy, spectral fidelity, partial-physics performance, and inference cost. Sec.~\ref{sec:discussion} analyzes the contributions of the individual components and discusses the limitations and future directions of the framework, followed by conclusions in Sec.~\ref{sec:conclusion}.

\section{Methodology}
\label{sec:method}

\begin{figure}[!t]
    \centering
    \begin{subfigure}[t]{\linewidth}
        \centering
        \includegraphics[width=0.95\linewidth]{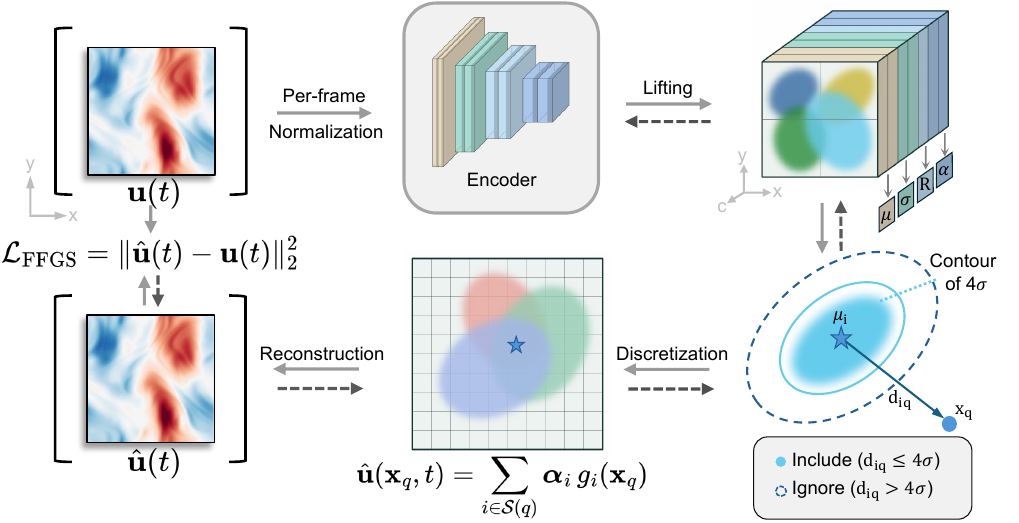}
        \caption{FFGS representation and local Gaussian rendering (Stage~1).}
        \label{fig:architecture_ffgs}
    \end{subfigure}
    \vspace{0.75em}
    \begin{subfigure}[t]{\linewidth}
        \centering
        \includegraphics[width=1.0\linewidth]{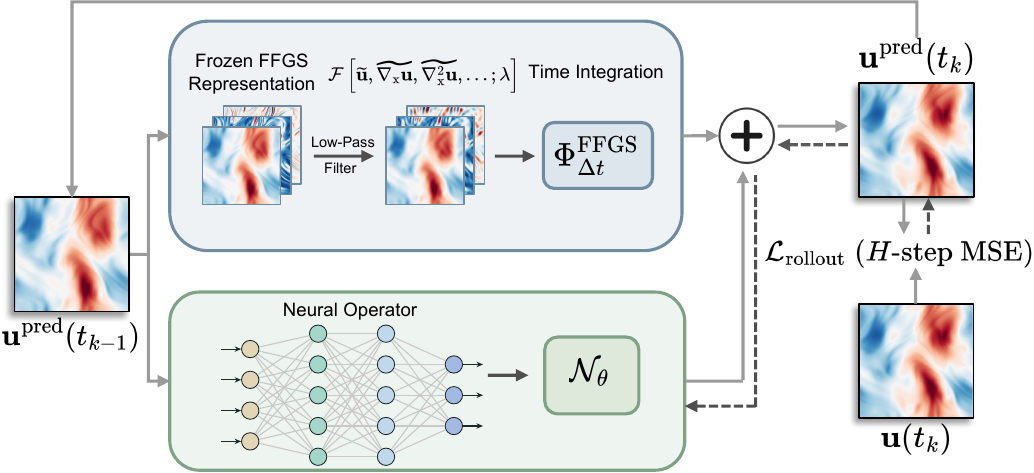}
        \caption{Physics-integrated neural operator (Stage~2).}
        \label{fig:architecture_overall}
    \end{subfigure}
    \caption{%
    Method overview. (a) The FFGS encoder maps a normalized grid field to anisotropic Gaussian primitives, which are rendered locally into a continuous field. The encoder is trained in Stage~1 by field reconstruction and then frozen. (b) The physics-integrated neural operator $\mathcal{G}^{\theta}_{\Delta t}$ adds the embedded-physics component $\Phi^{\mathrm{FFGS}}_{\Delta t}$ to the neural-operator component $\mathcal{N}_\theta$, with both acting on the same state. $\Phi^{\mathrm{FFGS}}_{\Delta t}$ evaluates the available governing terms with the filtered FFGS quantities and advances the state over $\Delta t$. Each prediction is fed back as the next input, starting from $\mathbf{u}^{\mathrm{pred}}(t_0)=\mathbf{u}(t_0)$. In Stage~2, both components are evaluated during the multi-step rollout, but only $\mathcal{N}_\theta$ is trained, and no gradients propagate through $\Phi^{\mathrm{FFGS}}_{\Delta t}$.}
    \label{fig:architecture}
\end{figure}
\subsection{Overview}
\label{sec:method:problem}

We consider a general class of nonlinear, coupled PDE systems for a $C$-component field in a parametric setting,
\begin{equation}
    \begin{aligned}
    \frac{\partial \mathbf{u}}{\partial t}
    &=
    \mathcal{F}\!\left[
        \mathbf{u},
        \nabla_{\mathbf{x}}\mathbf{u},
        \nabla_{\mathbf{x}}^{2}\mathbf{u},
        \ldots;
        \boldsymbol{\lambda}
    \right]  \\
    \mathcal{I}[\mathbf{u},\mathbf{u}_0]
    &= \mathbf{0},
    \qquad
    t=0,\ \mathbf{x}\in\Omega, \\
    \mathcal{B}[\mathbf{u},\nabla_{\mathbf{x}}\mathbf{u},\ldots]
    &= \mathbf{0},
    \qquad
    \mathbf{x}\in\partial\Omega ,
    \end{aligned}
    \label{eq:governing_pde}
\end{equation}
where $\mathbf{u}(\mathbf{x},t)\in\mathbb{R}^{C}$ is the solution field over the spatial domain $\Omega\subset\mathbb{R}^{d}$ and temporal horizon $t\in[0,T]$; $\mathcal{F}[\cdot;\boldsymbol{\lambda}]$ is a nonlinear operator acting on the field and its spatial derivatives, with $\boldsymbol{\lambda}$ denoting the physical parameters; and $\mathcal{I}$ and $\mathcal{B}$ encode the initial and boundary conditions, respectively. When the spatial argument is omitted, $\mathbf{u}(t)$ denotes the full field $\mathbf{u}(\cdot,t)$.

The governing PDE induces a finite-time solution operator $\mathcal{G}_{\Delta t}$, which maps the state $\mathbf{u}(t)$ to $\mathbf{u}(t+\Delta t)$ through the dynamics generated by $\mathcal{F}$. Specifically,
\begin{equation}
    \begin{aligned}
    \mathbf{u}(t+\Delta t)
    &=
    \mathcal{G}_{\Delta t}
    \!\left(
        \mathbf{u}(t);
        \boldsymbol{\lambda}
    \right) \\
    &=
    \mathbf{u}(t)
    +
    \int_{t}^{t+\Delta t}
    \mathcal{F}\!\left[
        \mathbf{u}(\tau),
        \nabla_{\mathbf{x}}\mathbf{u}(\tau),
        \nabla_{\mathbf{x}}^{2}\mathbf{u}(\tau),
        \ldots;
        \boldsymbol{\lambda}
    \right]
    \,d\tau .
    \end{aligned}
    \label{eq:flow_map}
\end{equation}

Numerical evaluation of this evolution generally requires temporal integration with a time step $\delta t$ constrained by the stability and accuracy requirements of the underlying integration scheme. Neural operators instead learn the finite-time solution operator $\mathcal{G}_{\Delta t}$ directly from data with a parameterized operator $\mathcal{N}_\theta$, so that a single evaluation can span multiple numerical time steps of size $\Delta t$.

To integrate the available physics into the neural operator through a continuous field representation, we construct an embedded-physics operator from the governing dynamics. Specifically, the operator $\operatorname{FFGS}$ reconstructs $\mathbf{u}(t)$ as a continuous field whose closed-form spatial derivatives provide the quantities required to evaluate the governing operator $\mathcal{F}$. Advancing $\mathbf{u}(t)$ over $\Delta t$ with the resulting governing terms defines the embedded-physics operator $\Phi_{\Delta t}^{\mathrm{FFGS}}(\mathbf{u}(t);\boldsymbol{\lambda})$. Embedding this operator in the forward map of $\mathcal{N}_\theta$ yields the physics-integrated neural operator $\mathcal{G}^{\theta}_{\Delta t}$,
\begin{equation}
  \mathbf{u}(t+\Delta t) = \mathcal{G}_{\Delta t}(\mathbf{u}(t);\boldsymbol{\lambda})
  \approx \mathcal{G}^{\theta}_{\Delta t}(\mathbf{u}(t);\boldsymbol{\lambda})
  = \underbrace{\mathcal{N}_\theta(\mathbf{u}(t);\boldsymbol{\lambda})}_{\substack{\text{neural-operator}\\ \text{component}}}
  + \underbrace{\Phi^{\mathrm{FFGS}}_{\Delta t}(\mathbf{u}(t);\boldsymbol{\lambda})}_{\substack{\text{embedded-physics}\\ \text{component}}} .
  \label{eq:hybrid}
\end{equation}

With the composite map in Eq.~\eqref{eq:hybrid} established, the following sections first construct $\Phi_{\Delta t}^{\mathrm{FFGS}}$ from the FFGS representation and its closed-form spatial derivatives, then describe the rollout training of $\mathcal{G}^{\theta}_{\Delta t}$, and finally extend the formulation to partially known physics.

\subsection{FFGS differentiable field representation}
\label{sec:method:ffgs}
We represent the state $\mathbf{u}(t)$ with a set of $G$ anisotropic Gaussian primitives $\{(\boldsymbol{\mu}_i, R_i, \boldsymbol{\sigma}_i, \boldsymbol{\alpha}_i)\}_{i=1}^{G}$. Here $\boldsymbol{\mu}_i\in\Omega$ is the center of the $i$-th primitive, $R_i$ its rotation matrix, $\boldsymbol{\sigma}_i$ its principal-axis scales, and $\boldsymbol{\alpha}_i$ its channel-wise amplitudes. The rotation is parameterized by an angle in 2D and by a unit quaternion in 3D. These parameters can be determined by minimizing the reconstruction error between the Gaussian representation and a given flow field. However, this approach requires repeating the optimization at every time step of an autoregressive rollout, which will introduce substantial computational overhead. FFGS avoids this issue by amortizing the fitting into an encoder $\mathcal{E}_\varphi$, which maps the state to the Gaussian parameters in a single forward pass. Further details on the encoder parameterization $\varphi$ are provided in~\ref{app:ffgs_param}. 

The predicted primitives reconstruct the field as an amplitude-weighted sum of Gaussian kernels,
\begin{equation}
  g_i(\mathbf{x}) = \exp\!\left[-\tfrac{1}{2}(\mathbf{x}-\boldsymbol{\mu}_i)^{\top}\Sigma_i^{-1}(\mathbf{x}-\boldsymbol{\mu}_i)\right],
  \qquad
  \Sigma_i = R_i\,\mathrm{diag}(\boldsymbol{\sigma}_i^{2})\,R_i^{\top},
  \label{eq:ffgs_cov}
\end{equation}
where the covariance $\Sigma_i$ determines its orientation and spatial extent. To avoid evaluating all $G$ kernels at every query point, FFGS exploits their rapid spatial decay and restricts the reconstruction to a fixed window of neighboring anchor cells,
\begin{equation}
\hat{\mathbf{u}}(\mathbf{x}_q,t) = \sum_{i\in\mathcal{S}(q)} \boldsymbol{\alpha}_ig_i(\mathbf{x}_q),
\label{eq:ffgs_render}
\end{equation}
where $\mathcal{S}(q)$ denotes the set of primitives associated with these cells. We discuss the window size, periodic boundary treatment, and resulting truncation error in~\ref{app:ffgs_render}.


Before rollout training, the encoder $\mathcal{E}_\varphi$ is trained separately by reconstructing field snapshots on the grid (Stage~1 in Fig.~\ref{fig:architecture}),
\begin{equation}
  \begin{aligned}
    \varphi^{*} &= \arg\min_{\varphi}\,\mathcal{L}_{\mathrm{FFGS}}(\varphi),\\
    \mathcal{L}_{\mathrm{FFGS}}(\varphi)
    &= \frac{1}{|\mathcal{M}|}\sum_{\mathbf{u}(t)\in\mathcal{M}}
       \frac{1}{N^{d}C}\sum_{j=1}^{N^{d}}
       \big\|\hat{\mathbf{u}}(\mathbf{x}_j,t)-\mathbf{u}(\mathbf{x}_j,t)\big\|_2^{2},
  \end{aligned}
  \label{eq:ffgs_loss}
\end{equation}
where $\mathcal{M}$ is the set of training snapshots, $\{\mathbf{x}_j\}_{j=1}^{N^{d}}$ are the grid points. The encoder is then frozen at $\varphi^{*}$.

With the encoder frozen, differentiating Eq.~\eqref{eq:ffgs_render} term by term gives the spatial derivatives of the reconstructed field in closed form. For each primitive,
\begin{equation}
  \begin{aligned}
    \nabla_{\mathbf{x}} g_i(\mathbf{x}) &= -g_i(\mathbf{x})\,\Sigma_i^{-1}(\mathbf{x}-\boldsymbol{\mu}_i),\\
    \nabla_{\mathbf{x}}^{2} g_i(\mathbf{x}) &= g_i(\mathbf{x})\left[(\mathbf{x}-\boldsymbol{\mu}_i)^{\top}\Sigma_i^{-2}(\mathbf{x}-\boldsymbol{\mu}_i)-\operatorname{tr}\!\left(\Sigma_i^{-1}\right)\right],
  \end{aligned}
  \label{eq:ffgs_kernel_deriv}
\end{equation}
and summing over $\mathcal{S}(q)$ gives the derivatives of each channel of the reconstructed field,
\begin{equation}
  \begin{aligned}
    \nabla_{\mathbf{x}} \hat{u}_c(\mathbf{x}_q,t) &= \sum_{i\in\mathcal{S}(q)} \alpha_{i,c}\,\nabla_{\mathbf{x}} g_i(\mathbf{x}_q),\\
    \nabla_{\mathbf{x}}^{2} \hat{u}_c(\mathbf{x}_q,t) &= \sum_{i\in\mathcal{S}(q)} \alpha_{i,c}\,\nabla_{\mathbf{x}}^{2} g_i(\mathbf{x}_q), \qquad c=1,\dots,C.
  \end{aligned}
  \label{eq:ffgs_field_deriv}
\end{equation}
The derivatives thus require neither finite-difference stencils nor automatic differentiation with respect to the spatial coordinates.

These derivatives are exact for the reconstructed field, but differentiation amplifies the high-wavenumber errors of the reconstruction. We therefore apply a low-pass filter $\Pi_{\kappa_c}$ to the reconstructed field and its derivatives before they enter the governing operator,
\begin{equation}
  \tilde{\mathbf{u}}(t) = \Pi_{\kappa_c} \hat{\mathbf{u}}(t),
  \quad
  \widetilde{\nabla_{\mathbf{x}}\mathbf{u}}(t) = \Pi_{\kappa_c} \nabla_{\mathbf{x}}\hat{\mathbf{u}}(t),
  \quad
  \widetilde{\nabla^{2}_{\mathbf{x}}\mathbf{u}}(t) = \Pi_{\kappa_c} \nabla^{2}_{\mathbf{x}}\hat{\mathbf{u}}(t),
  \label{eq:ffgs_filter}
\end{equation}
where $\Pi_{\kappa_c}$ acts in Fourier space on the grid samples of each channel and attenuates wavenumbers above the cutoff $\kappa_c$. 

When the available terms of $\mathcal{F}$ are evaluated, terms with spatial derivatives are computed from Eq.~\eqref{eq:ffgs_filter} and all other terms directly from the grid state. The embedded-physics operator $\Phi_{\Delta t}^{\mathrm{FFGS}}$ in Eq.~\eqref{eq:hybrid} is then obtained by advancing $\mathbf{u}(t)$ over $\Delta t$ with the resulting right-hand side and restricting the updated state to the band retained by $\Pi_{\kappa_c}$. The filter form, the benchmark-specific cutoffs, and the time-integration scheme are detailed in~\ref{app:filter_rk4}.

\subsection{Temporal Evolution}
\label{sec:method:train}

With the encoder fixed at $\varphi^{*}$, Stage~2 (Fig.~\ref{fig:architecture}) trains the physics-integrated neural operator $\mathcal{G}^{\theta}_{\Delta t}$. Both of its components are evaluated at every step of this stage, but only the parameters $\theta$ of $\mathcal{N}_\theta$ are updated. These parameters minimize the prediction error over a rollout of $H$ steps,
\begin{equation}
  \begin{aligned}
    \theta^{*} &= \arg\min_{\theta}\,\mathcal{L}_{\mathrm{rollout}}(\theta),\\
    \mathcal{L}_{\mathrm{rollout}}(\theta)
    &= \frac{1}{H}\sum_{k=1}^{H}\frac{1}{N^{d}C}\sum_{j=1}^{N^{d}}
       \big\|\mathbf{u}^{\mathrm{pred}}(\mathbf{x}_j,t_k)-\mathbf{u}(\mathbf{x}_j,t_k)\big\|_2^{2},
  \end{aligned}
  \label{eq:rollout_loss}
\end{equation}
where the predictions are generated by applying Eq.~\eqref{eq:hybrid} recursively from a ground-truth state $\mathbf{u}(t_0)$,
\begin{equation}
  \mathbf{u}^{\mathrm{pred}}(t_k)
  = \mathcal{N}_\theta\big(\mathbf{u}^{\mathrm{pred}}(t_{k-1});\boldsymbol{\lambda}\big)
  + \Phi^{\mathrm{FFGS}}_{\Delta t}\big(\mathbf{u}^{\mathrm{pred}}(t_{k-1});\boldsymbol{\lambda}\big),
  \qquad k=1,\dots,H,
  \label{eq:rollout}
\end{equation}
with $t_k=t_0+k\Delta t$ and $\mathbf{u}^{\mathrm{pred}}(t_0)=\mathbf{u}(t_0)$. Training on these recursively predicted states exposes $\mathcal{N}_\theta$ to the states generated by $\mathcal{G}^{\theta}_{\Delta t}$ itself, rather than restricting optimization to one-step transitions from ground-truth inputs. At inference, $\mathcal{G}^{\theta}_{\Delta t}$ is applied in the same autoregressive manner over the target horizon.

\subsection{Extension to partially known physics}
\label{sec:method:partial}
We next consider the partially known setting, where the governing form is known but some of its coefficients are unknown. Even with the governing form and trajectory data at hand, purely data-driven neural operators offer no natural way to identify these coefficients and incorporate the resulting physics into the model. However, FFGS makes this possible. Specifically, with the trained FFGS encoder, all spatial derivatives in the governing equation are available in closed form (Eq.~\eqref{eq:ffgs_field_deriv}). The time derivative $\partial_t\mathbf{u}$ is estimated from several observed frames. When $\mathcal{F}$ depends linearly on the unknown coefficients, these quantities allow the coefficients to be identified directly by least squares,
\begin{equation}
    \hat{\boldsymbol{\lambda}}=\arg\min_{\boldsymbol{\lambda}}\big\|\mathbf{A}\boldsymbol{\lambda}-\mathbf{b}\big\|_2^{2},
  \label{eq:coef_ls}
\end{equation}
where $\boldsymbol{\lambda}$ contains only the unknown coefficients, and each column of $\mathbf{A}$ is the term of $\mathcal{F}$ associated with one of the unknown coefficients, computed from the FFGS reconstruction. The vector $\mathbf{b}$ contains the estimated $\partial_t\mathbf{u}$ minus the terms of $\mathcal{F}$ with known coefficients. The estimate $\hat{\boldsymbol{\lambda}}$ is computed once for each trajectory from its first few frames (\ref{app:coef_id}). It then replaces $\boldsymbol{\lambda}$ in $\Phi^{\mathrm{FFGS}}_{\Delta t}$, so that the resulting physics is incorporated into the model. The same estimate is also supplied to $\mathcal{N}_\theta$ as a conditioning input.

\section{Experimental setup}
\label{sec:setup}

\subsection{Benchmark suite and data generation}

\label{sec:setup:benchmarks}
We evaluate the physics-integrated neural operator in two regimes. In the full-physics regime, the governing form and its coefficients are prescribed and shared by all trajectories. Five full-physics benchmarks cover pure transport, transport--diffusion, nonlinear self-advection, reaction-driven moving interfaces, and three-dimensional transport--diffusion (Table~\ref{tab:benchmarks}). In the partial-physics regime, the governing form is prescribed, but its coefficients vary across trajectories and are withheld. They are identified from the first few frames of each trajectory (Section~\ref{sec:method:partial}). 

\begin{table}[!t]
\centering
\small
\setlength{\tabcolsep}{4pt}
\renewcommand{\arraystretch}{1.15}
\begin{adjustbox}{max width=\textwidth}
\begin{tabular}{lllll}
\toprule
\textbf{Case} & \textbf{Governing equation} & \textbf{Coefficients}
& \textbf{Grid} & $\Delta t$ \\
\midrule
2D Adv& $\partial_t u = -\mathbf{v}\cdot\nabla u$& $\mathbf{v}=0.025\,\mathbf{1}$& $160^2$ & $1.0$ \\
2D Adv--Diff& $\partial_t u = -\mathbf{v}\cdot\nabla u + D\nabla^2 u$& $\mathbf{v}=0.025\,\mathbf{1},\ D=2{\times}10^{-6}$& $160^2$ & $1.0$ \\
2D Burgers& $\partial_t \mathbf{u} = -c(\mathbf{u}\cdot\nabla)\mathbf{u}
   + \nu\nabla^2\mathbf{u}$& $c=4.69{\times}10^{-3},\ \nu=10^{-4}$& $160^2$ & $1.0$ \\
2D Adv--Allen--Cahn& $\partial_t u = -\mathbf{v}\cdot\nabla u + \nu\nabla^2 u
   + r(u-u^3)$& $\mathbf{v}=0.05\,\mathbf{1},\ \nu=10^{-3},\ r=1$& $160^2$ & $0.1$ \\
3D Adv--Diff& $\partial_t u = -\mathbf{v}\cdot\nabla u + D\nabla^2 u$& $\mathbf{v}=0.0125\,\mathbf{1},\ D=10^{-6}$& $64^3$ & $1.0$ \\
\bottomrule
\end{tabular}
\end{adjustbox}
\vspace{0.35em}
\begin{minipage}{0.96\textwidth}
\footnotesize
All quantities are nondimensional, and all problems are posed on the unit periodic domain $[0,1)^d$. Initial conditions are sampled as random truncated Fourier series with modes restricted to $|k_i|\le k_{\max}=5$ along each spatial axis. Each full-physics benchmark contains 256 training trajectories and 30 held-out test trajectories. Models are trained on the first 50 steps and evaluated by 200-step autoregressive rollout. The velocity is prescribed as $\mathbf{v}=v\mathbf{1}$, where $\mathbf{1}$ is the vector of ones. The symbols $c$, $D$, $\nu$, and $r$ denote the convection scaling, diffusivity, viscosity, and reaction rate, respectively.
\end{minipage}
\caption{Full-physics benchmarks and data-generation parameters.}
\label{tab:benchmarks}
\end{table}

All trajectories are generated using the same pseudo-spectral numerical framework as APEBench~\cite{koehler_apebench_2024}, with governing equations and coefficients specified for the present experiments. Table~\ref{tab:benchmarks} lists the simulation parameters together with the training and evaluation horizons. The train and test splits are fixed for each benchmark and shared by all methods. This protocol evaluates generalization to unseen initial conditions and temporal extrapolation beyond the training horizon.

The partial-physics benchmark uses the two-dimensional advection--diffusion equation of Table~\ref{tab:benchmarks}, with $\boldsymbol{\lambda}=(v,D)$ sampled log-uniformly per trajectory within the ranges given in~\ref{app:partial_data}. Its training and test sets contain 512 and 30 trajectories, with the same horizons as the full-physics benchmarks. The first $W=7$ frames of each test trajectory are provided to all methods as observed history, and evaluation starts from frame $W$.

\subsection{Models and baselines}
\label{sec:setup:model_training}
\label{sec:setup:baselines}

The FFGS encoder uses an anchor lattice with $A=20$ cells per dimension and a rendering window of half-width $W_{\mathrm{loc}}=4$ anchor cells. The neural-operator component $\mathcal{N}_\theta$ is an FNO, trained with a rollout length of $H=5$. In the partial-physics regime, it is conditioned on $\hat{\boldsymbol{\lambda}}$ through feature-wise linear modulation (FiLM).

The purely data-driven baselines are FNO~\cite{li_fourier_2021}, U-Net~\cite{ronneberger_unet_2015}, and ResNet~\cite{he_resnet_2016}, trained with the same input--output signature and protocol as the proposed model. In the partial-physics regime, they instead take the most recent $W$ frames as input. Parameter budgets are matched at the level of the trainable dynamics map, which for the proposed model is $\mathcal{N}_\theta$. The frozen FFGS encoder is reported separately in~\ref{app:ffgs_encoder}. Variants of the baselines with tenfold trainable budgets are additionally evaluated on the Burgers and partial-physics benchmarks (\ref{app:supp_capacity}) and partial-physics (Section~\ref{sec:partial}) benchmarks. Architectures and hyperparameters of all networks are given in~\ref{app:no_impl}.

\subsection{Evaluation metrics}
\label{sec:setup:metrics}
The field-level metrics are built on the mean squared error $\mathrm{MSE}(k)$ at lead time $k$. The primary metric is the relative $\ell_2$ error $\mathrm{r}\ell_2(k)$, which normalizes $\mathrm{MSE}(k)$ by the mean square of the reference field,
\begin{equation}
\begin{aligned}
\mathrm{MSE}(k)
&=\frac{1}{N^{d}C}\sum_{j=1}^{N^{d}}
\big\|\mathbf{u}^{\mathrm{pred}}(\mathbf{x}_j,t_k)-\mathbf{u}(\mathbf{x}_j,t_k)\big\|_2^{2},\\
\mathrm{r}\ell_2(k)
&=\left(\frac{\mathrm{MSE}(k)}{\frac{1}{N^{d}C}\sum_{j=1}^{N^{d}}\big\|\mathbf{u}(\mathbf{x}_j,t_k)\big\|_2^{2}}\right)^{1/2},
\end{aligned}
\qquad k=1,\dots,T_{\mathrm{roll}},
\label{eq:rel_l2}
\end{equation}
where $t_k=t_0+k\Delta t$ and $t_0$ is the initial time of the evaluated rollout. The evaluated rollout has $T_{\mathrm{roll}}=200$ steps in the full-physics regime and $T_{\mathrm{roll}}=194$ in the partial-physics regime. For rollout curves, $\mathrm{r}\ell_2(k)$ is averaged over the $N_{\mathrm{test}}=30$ held-out test trajectories at each lead time. For the summary tables, the error is first averaged over the evaluated rollout of each trajectory and is then reported as the mean $\pm$ one standard deviation across the 30 trajectory-level averages.

To assess spectral fidelity, we compute the radial energy spectrum of each rollout frame. For a field $\mathbf{u}(\mathbf{x})$ with channel-wise Fourier coefficients $U_c(\mathbf{q})$, the radial spectrum at integer wavenumber $\kappa$ is defined as
\begin{equation}
E(\kappa)
=
\sum_{
\kappa-\frac{1}{2}
\le
|\mathbf{q}|
<
\kappa+\frac{1}{2}
}
\;
\sum_{c=1}^{C}
\left|
U_c(\mathbf{q})
\right|^2,
\label{eq:radial_spectrum}
\end{equation}
where $\mathbf{q}$ denotes the spatial Fourier mode, and the channel sum is omitted for scalar fields.

The radial spectrum is then averaged over all rollout frames and held-out test trajectories,
\begin{equation}
\overline{E}(\kappa)
=
\frac{1}
{N_{\mathrm{test}}T_{\mathrm{roll}}}
\sum_{n=1}^{N_{\mathrm{test}}}
\sum_{k=1}^{T_{\mathrm{roll}}}
E_{n,k}(\kappa),
\label{eq:mean_radial_spectrum}
\end{equation}
where $E_{n,k}$ is the spectrum of trajectory $n$ at $t_k$. The spectral error is then defined as
\begin{equation}
\mathrm{Spec.\,err}
=
\frac{
\left\|
\overline{E}^{\mathrm{pred}}
-
\overline{E}^{\mathrm{GT}}
\right\|_2
}{
\left\|
\overline{E}^{\mathrm{GT}}
\right\|_2
}.
\label{eq:spectral_metric}
\end{equation}

The $\kappa=0$ mode represents the domain mean rather than nonzero-wavenumber spectral content. For the advection, advection--diffusion, and Burgers benchmarks, the reference mean is identically zero and exactly conserved. Accordingly, Eq.~\eqref{eq:spectral_metric} is evaluated over $\kappa\ge1$ for these benchmarks. For the advection--Allen--Cahn benchmark, the order parameter has an evolving, non-conserved mean, and the $\kappa=0$ mode is therefore retained.

For completeness, we also report the peak signal-to-noise ratio,
\begin{equation}
\mathrm{PSNR}(k)=10\log_{10}\!\left(\frac{R^{2}}{\mathrm{MSE}(k)}\right),
\label{eq:psnr}
\end{equation}
where $R$ is the global dynamic range of the reference data over the held-out test set. PSNR is aggregated in the same way as $\mathrm{r}\ell_2$. Lower $\mathrm{r}\ell_2$ and spectral error, and higher PSNR, indicate better performance.

\section{Results}
\label{sec:results}

\subsection{Aggregate rollout accuracy}
\label{sec:quant}

Table~\ref{tab:metrics_comparison} summarizes aggregate test performance over 200-step autoregressive rollouts on the five full-physics benchmarks. The proposed physics-integrated neural operator achieves the best performance on all benchmarks and all reported metrics. Compared with the strongest purely data-driven baseline in each benchmark, selected by mean $\mathrm{r}\ell_2$ error, it reduces $\mathrm{r}\ell_2$ by factors ranging from $1.5\times$ on 3D advection--diffusion to $2.2\times$ on Burgers. The corresponding PSNR gains range from $3.3$ to $6.6$~dB.
\begin{table*}[!t]
\centering
\resizebox{\textwidth}{!}{%
\begin{tabular}{llcccc}
\toprule
\textbf{Benchmark} & \textbf{Method} & \textbf{\#Params}
& \textbf{PSNR$\uparrow$ (dB)} & \textbf{$\mathrm{r}\ell_2\downarrow$}
& \textbf{Spec.\,err$\downarrow$} \\
\midrule
\multirow{4}{*}{Advection (2D)}
 & FNO & 57.8k & $40.04\pm1.68$ & $0.089\pm0.011$ & $0.0214$ \\
 & U-Net & 55.7k & $36.20\pm1.14$ & $0.129\pm0.025$ & $0.0403$ \\
 & ResNet & 61.2k & $29.42\pm0.66$ & $0.394\pm0.028$ & $0.2898$ \\
 & \textbf{Ours} & 57.8k & $\mathbf{44.75\pm1.92}$ & $\mathbf{0.052\pm0.014}$ & $\mathbf{0.0031}$ \\
\midrule
\multirow{4}{*}{Adv--Diffusion (2D)}
 & FNO & 57.8k & $41.41\pm1.75$ & $0.084\pm0.011$ & $0.0276$ \\
 & U-Net & 55.7k & $36.37\pm1.70$ & $0.146\pm0.037$ & $0.0894$ \\
 & ResNet & 61.2k & $33.89\pm0.85$ & $0.207\pm0.014$ & $0.0990$ \\
 & \textbf{Ours} & 57.8k & $\mathbf{46.02\pm1.83}$ & $\mathbf{0.050\pm0.011}$ & $\mathbf{0.0067}$ \\
\midrule
\multirow{4}{*}{Burgers (2D)}
 & FNO & 57.8k & $43.03\pm1.86$ & $0.236\pm0.047$ & $0.1102$ \\
 & U-Net & 55.8k & $46.07\pm2.02$ & $0.225\pm0.096$ & $0.1489$ \\
 & ResNet & 61.2k & $49.83\pm0.82$ & $0.127\pm0.034$ & $0.0259$ \\
 & \textbf{Ours} & 57.8k & $\mathbf{56.40\pm1.77}$ & $\mathbf{0.057\pm0.023}$ & $\mathbf{0.0007}$ \\
\midrule
\multirow{4}{*}{Adv--Allen--Cahn (2D)}
 & FNO & 57.8k & $20.74\pm3.23$ & $0.313\pm0.151$ & $0.1086$ \\
 & U-Net & 55.7k & $35.13\pm4.23$ & $0.062\pm0.048$ & $0.0367$ \\
 & ResNet & 61.2k & $46.77\pm2.01$ & $0.014\pm0.005$ & $0.0033$ \\
 & \textbf{Ours} & 57.8k & $\mathbf{51.36\pm2.97}$ & $\mathbf{0.008\pm0.004}$ & $\mathbf{0.0010}$ \\
\midrule
\multirow{4}{*}{Adv--Diffusion (3D)}
 & FNO & 1.15M & $37.82\pm0.61$ & $0.157\pm0.009$ & $0.0715$ \\
 & U-Net & 1.12M & $41.40\pm0.97$ & $0.100\pm0.010$ & $0.0173$ \\
 & ResNet & 1.15M & $30.81\pm0.46$ & $0.343\pm0.027$ & $0.2206$ \\
 & \textbf{Ours} & 1.15M & $\mathbf{44.67\pm0.82}$ & $\mathbf{0.068\pm0.005}$ & $\mathbf{0.0021}$ \\
\bottomrule
\end{tabular}}
\vspace{0.35em}

\begin{minipage}{0.98\textwidth}
\footnotesize
Reporting conventions follow Section~\ref{sec:setup:metrics}. \#Params counts the trainable parameters of the learned dynamics map. The FFGS encoder is trained separately for representation reconstruction and frozen during rollout training. Bold entries mark the best value within each benchmark.
\end{minipage}

\caption{Aggregate test-rollout performance on the five full-physics PDE benchmarks.}
\label{tab:metrics_comparison}
\end{table*}

Among the baselines, FNO performs best on the 2D linear transport cases, ResNet on the nonlinear cases, and U-Net on 3D advection--diffusion. No single baseline architecture is best across all benchmarks, as also observed in the architecture comparison of APEBench~\cite{koehler_apebench_2024}. FNO is the most direct comparison, as the neural-operator component of the proposed model has the same input--output signature and trainable dynamics-map budget. The gain over FNO is therefore consistent with a contribution from the embedded-physics component, which Section~\ref{sec:ablations} examines further. Relative to the strongest baseline of each benchmark, the spectral error decreases by larger factors than $\mathrm{r}\ell_2$, from $3.3\times$ on advection--Allen--Cahn to more than $30\times$ on Burgers. Section~\ref{sec:spectral} examines this metric in more detail.
\subsection{Error evolution over the rollout}
\label{sec:rollout}

\begin{figure}[!t]
    \centering
    \includegraphics[width=\linewidth]{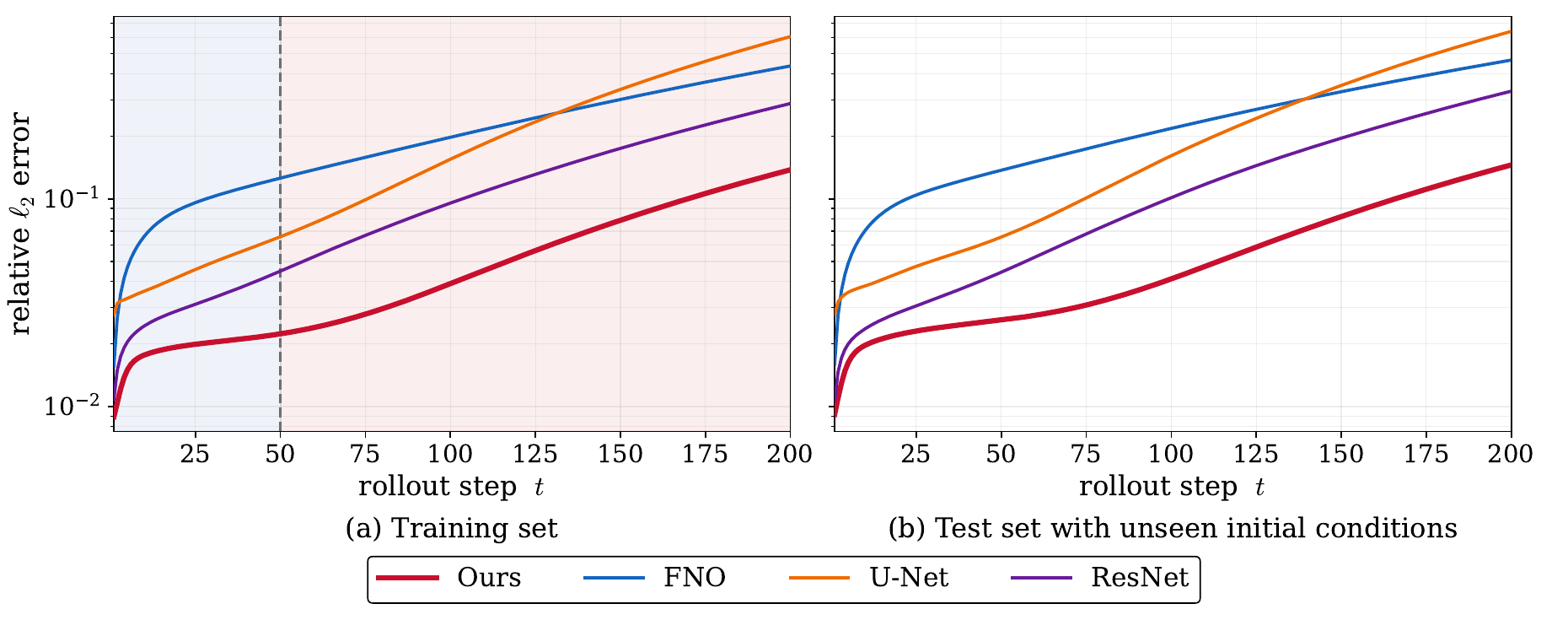}
    \caption{%
    Error evolution over the rollout on the Burgers benchmark. Relative $\ell_2$ error is plotted on a logarithmic scale for the proposed surrogate and three purely data-driven baselines. Panel (a) shows rollouts on the training trajectories, and panel (b) shows rollouts on held-out trajectories with unseen initial conditions. The shaded region in panel (a) indicates extrapolation beyond the 50-step training horizon ($t>50$).}
    \label{fig:rollout}
\end{figure}

Figure~\ref{fig:rollout} shows how the $\mathrm{r}\ell_2$ error evolves over the rollout on the Burgers benchmark. The rollout evaluates both extrapolation beyond the 50-step training horizon and generalization to unseen initial conditions. The proposed model attains the lowest $\mathrm{r}\ell_2$ at every step, on both the training and held-out trajectories. Within the training horizon, its error remains near $2\times10^{-2}$, whereas the baseline errors grow to between $4.5\times10^{-2}$ and $1.2\times10^{-1}$. FNO shows the fastest initial growth and reaches about $10^{-1}$ within the first 25 steps. Beyond the horizon, all errors increase, and U-Net degrades fastest, overtaking FNO after roughly 130 steps. The error of the proposed model grows at a rate similar to that of ResNet, the baseline with the lowest mean $\mathrm{r}\ell_2$. Its margin of roughly a factor of two therefore persists up to $t=200$, in line with the $2.2\times$ reduction in mean $\mathrm{r}\ell_2$ reported in Table~\ref{tab:metrics_comparison}. The training and held-out panels are nearly identical. The embedded-physics component evaluates the available governing terms at every step, including steps beyond the training horizon and trajectories with unseen initial conditions. This is consistent with the advantage of the proposed model persisting in both settings. Rollout curves for the other four benchmarks are given in~\ref{app:supp_rollout}.
\subsection{Field-level comparison}
\label{sec:field3d}

\begin{figure*}[!t]
    \centering
    \includegraphics[width=\textwidth]{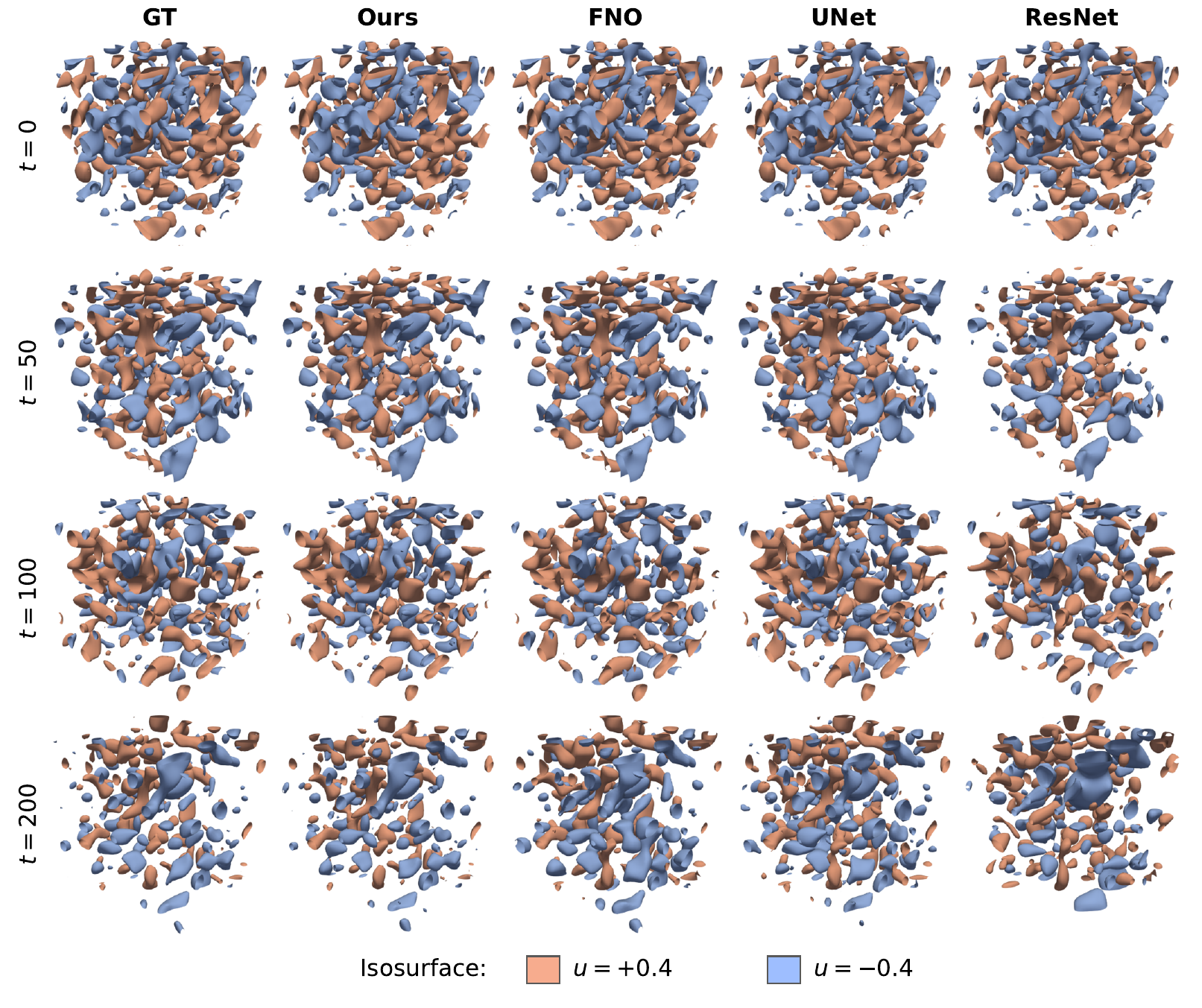}
    \caption{Three-dimensional advection--diffusion rollout for a representative held-out trajectory at large P\'eclet number. Positive and negative scalar structures are visualized using the isosurfaces $u=+0.4$ and $u=-0.4$, respectively. Columns correspond to the ground truth, the proposed model, FNO, U-Net, and ResNet. Rows show the initial condition and the autoregressive predictions at $t=50$, $100$, and $200$. All methods are initialized from the same ground-truth field at $t=0$, and the viewing direction and isosurface levels are fixed across all panels.}
    \label{fig:advdiff3d_field}
\end{figure*}

Figure~\ref{fig:advdiff3d_field} compares the evolution of the three-dimensional scalar field over a long autoregressive rollout. The paired isosurfaces show both the transport of the field and the deformation of its positive and negative structures. These features may not be apparent from domain-averaged error metrics.

At $t=50$, all methods reproduce the dominant transported structures, but the baselines already deviate in the geometry and placement of smaller components, most visibly for ResNet. These deviations grow by $t=100$, where FNO, U-Net, and particularly ResNet show deformation, merging, or disappearance of individual structures. The proposed model stays closer to the ground truth in the spatial organization, sign, and characteristic length scales of the structures.

By $t=200$, the baselines deviate markedly from the ground truth, whereas the proposed model still retains the principal positive and negative structures with the least distortion of their topology and spatial distribution. FNO and U-Net preserve parts of the large-scale organization but with larger displacement and morphological deviation. ResNet departs most strongly from the reference, including a large merged negative structure near the top of the domain. This ordering matches the three-dimensional results in Table~\ref{tab:metrics_comparison}.

\subsection{Spectral fidelity}
\label{sec:spectral}

\begin{figure*}[!t]
    \centering
    \includegraphics[width=\linewidth]{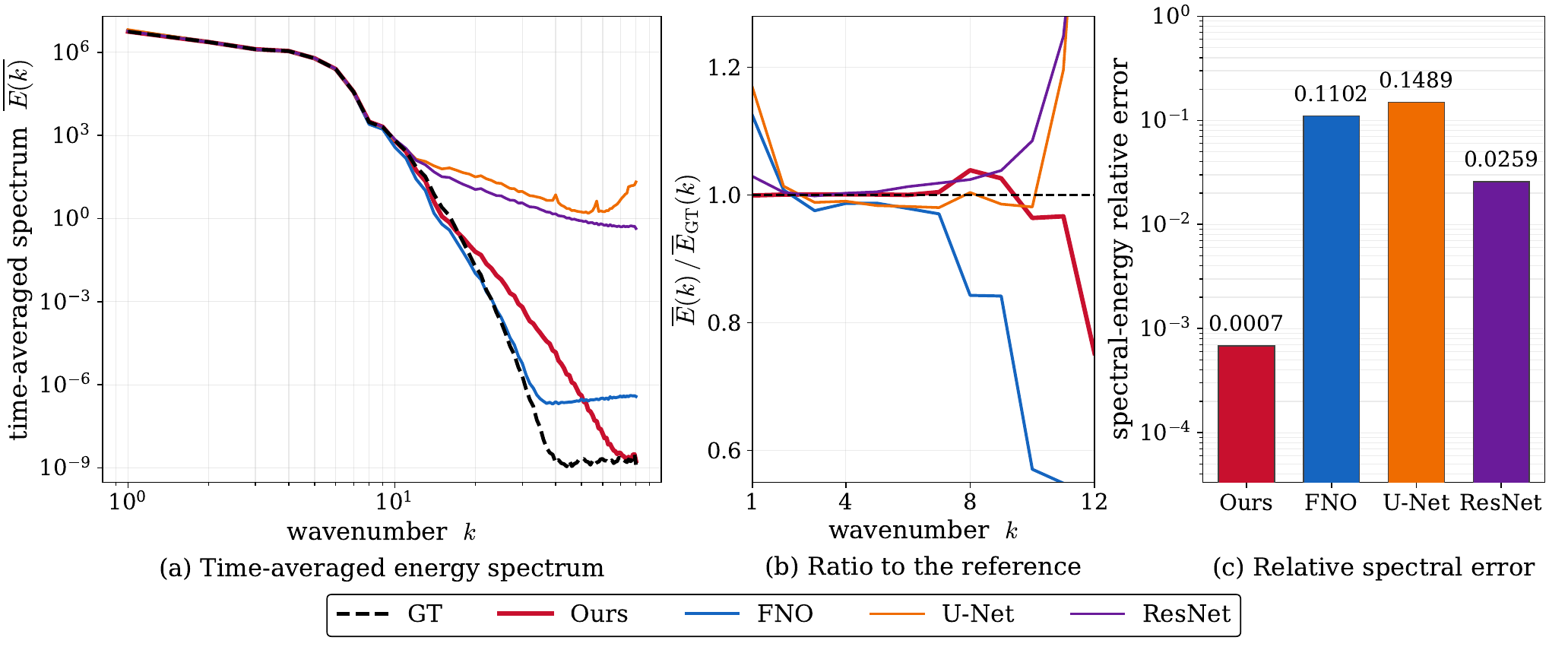}
    \caption{%
    Spectral fidelity on the 2D Burgers test set
    ($\nu=10^{-4}$; 30 held-out trajectories $\times$ 200 rollout steps).
    (a) Time-averaged radial energy spectrum
    $\overline{E}(\kappa)$
    for the reference solution and each model.
    (b) Ratio
    $\overline{E}(\kappa)/
    \overline{E}_{\mathrm{GT}}(\kappa)$
    over the energy-containing range.
    (c) Relative error of the time-averaged spectrum,
    $\|\overline{E}-\overline{E}_{\mathrm{GT}}\|_2/
    \|\overline{E}_{\mathrm{GT}}\|_2$,
    evaluated over $\kappa\ge1$
    (Section~\ref{sec:setup:metrics}).
    }
  \label{fig:burgers2d_spectrum_all}
\end{figure*}

Figure~\ref{fig:burgers2d_spectrum_all} compares the time-averaged radial energy spectra of the 2D Burgers rollouts. As shown in Fig.~\ref{fig:burgers2d_spectrum_all}(a), all methods closely follow the reference spectrum up to $\kappa\approx10$. At higher wavenumbers, U-Net and ResNet accumulate spurious energy and level off at $10^{0}$ to $10^{1}$, whereas the reference decays to about $10^{-9}$. FNO follows the reference most closely through this decay and then levels off at a few times $10^{-7}$. The proposed model decays more slowly but reaches the reference level at the highest wavenumbers.

The spectral ratio in Fig.~\ref{fig:burgers2d_spectrum_all}(b) shows that FNO, U-Net, and ResNet overestimate the energy at $\kappa=1$ by about 13, 17, and 3 percent, respectively, whereas the proposed model remains nearly indistinguishable from the reference. This overestimation at the most energetic wavenumber dominates the spectral error in Fig.~\ref{fig:burgers2d_spectrum_all}(c). The spectral error is $0.0007$ for the proposed model, compared with $0.1102$ for FNO, $0.1489$ for U-Net, and $0.0259$ for ResNet. The improved spectral fidelity is consistent with the proposed physics-integrated formulation. Spectra for the other four benchmarks are given in~\ref{app:supp_spectra}.

\subsection{Partial-physics regime}
\label{sec:partial}

\begin{figure*}[!t]
    \centering
    \includegraphics[width=0.90\linewidth]{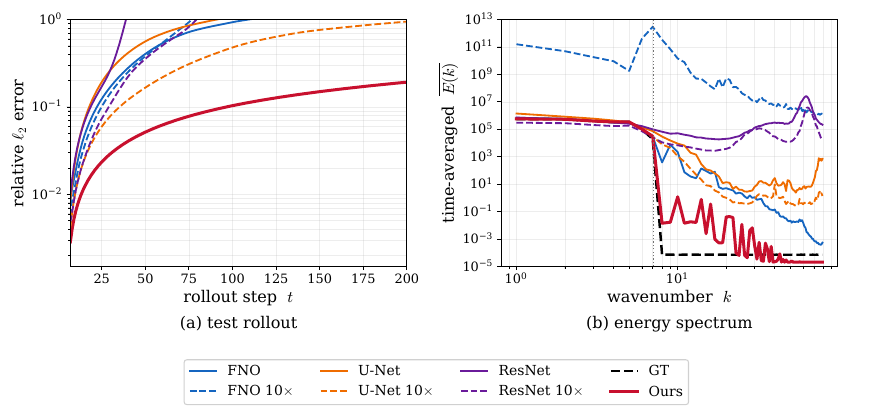}
    \caption{Partial-physics regime on the advection--diffusion benchmark (30 held-out trajectories $\times$ 194 rollout steps). (a) $\mathrm{r}\ell_2$ error of the proposed model and the baselines at matched and tenfold ($10\times$) trainable dynamics-map budgets. (b) Time-averaged energy spectra. The dotted line marks the band edge of the initial conditions.}
    \label{fig:partial_baselines}
\end{figure*}

Using the closed-form FFGS derivatives, the regression in Eq.~\eqref{eq:coef_ls} identifies the coefficients $\boldsymbol{\lambda}=(v,D)$ of each trajectory from only its $W=7$ observed frames (\ref{app:coef_id}). Over the 30 test trajectories, the mean relative error of the identified coefficients is $0.34\%$ for $v$ and $2.84\%$ for $D$. The larger error in $D$ is consistent with the large P\'eclet number of this benchmark, at which the diffusive term contributes little to $\partial_t u$.

Figure~\ref{fig:partial_baselines}(a) compares the $\mathrm{r}\ell_2$ error of the proposed model with that of the baselines. The proposed model has the lowest error throughout the rollout and reaches $0.19$ at $t=200$. In contrast, the errors of all matched-budget baselines reach $10^{0}$ between $t\approx40$ and $t\approx105$. Even with a tenfold trainable budget, the most accurate baseline, U-Net~$10\times$, reaches about $0.9$ at $t=200$, nearly five times the error of the proposed model. This suggests that the advantage of the proposed model comes from its physics-integrated formulation rather than from model capacity.

For this linear equation, the reference spectrum drops to the numerical floor beyond the band edge of the initial conditions, marked by the dotted line in Fig.~\ref{fig:partial_baselines}(b). The proposed model largely preserves this band-limited structure. Its spectrum falls by about five orders of magnitude at the band edge, stays below $10^{0}$ beyond it, and reaches the floor beyond $\kappa\approx40$. The baselines instead accumulate spurious energy beyond the band edge, at least one order of magnitude more than the proposed model. FNO $10\times$ even exceeds the reference by about five orders of magnitude within the band. The preserved band-limited structure is consistent with the embedded-physics component, which evaluates the advection and diffusion terms with the identified coefficients at every step. The advantage observed in the full-physics regime thus persists when the coefficients are identified from data rather than prescribed.
 
\subsection{Inference cost}
\label{sec:cost}

\begin{figure*}[!t]
    \centering
    \includegraphics[width=\linewidth]{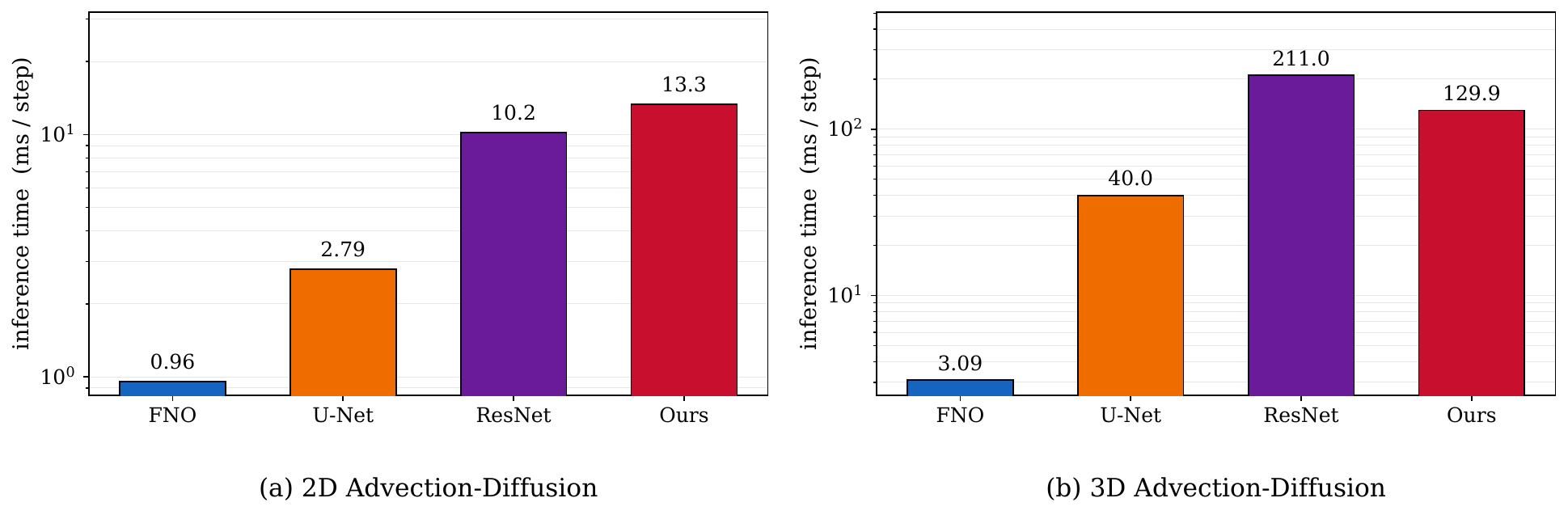}
    \caption{Per-step inference cost measured as wall-clock time per
    autoregressive update. Values are the minimum over five timed
    rollouts after warm-up and are shown on a logarithmic scale.}
    \label{fig:cost}
\end{figure*}

Figure~\ref{fig:cost} reports the wall-clock time per autoregressive step on 2D and 3D advection--diffusion. The proposed model requires 13.3~ms per step in 2D and 129.9~ms in 3D. It is slower than FNO and U-Net on both benchmarks and comparable to ResNet, which requires 10.2~ms in 2D and 211.0~ms in 3D. In 3D, the proposed model is faster than ResNet, and its $\mathrm{r}\ell_2$ error is one fifth of that of ResNet. Its neural-operator component shares the architecture of the FNO baseline. The difference from FNO, about 12~ms in 2D and 127~ms in 3D, therefore reflects the cost of the embedded-physics component. In return for this cost, the $\mathrm{r}\ell_2$ error decreases relative to FNO from 0.084 to 0.050 in 2D and from 0.157 to 0.068 in 3D (Table~\ref{tab:metrics_comparison}).


\section{Discussion}
\label{sec:discussion}

\subsection{Component analysis}
\label{sec:ablations}

\begin{figure*}[!t]
    \centering
    \includegraphics[width=\linewidth]{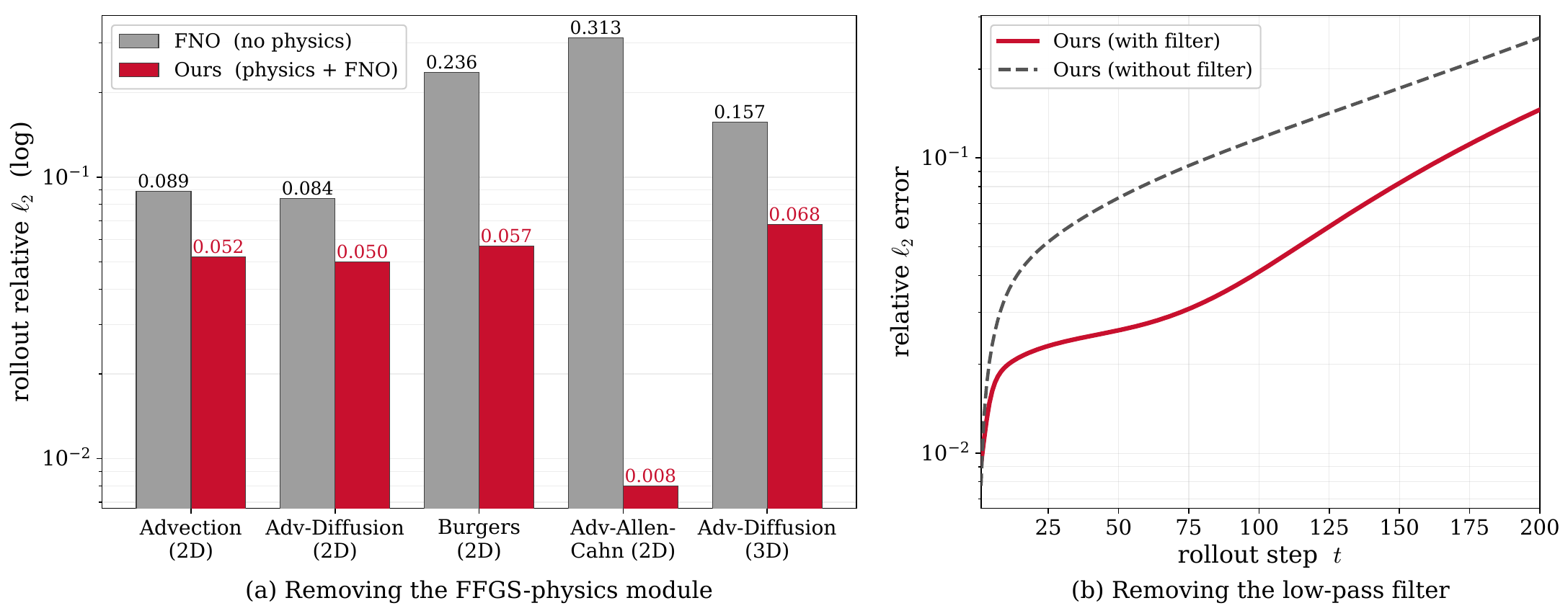}
    \caption{Ablation in the full-physics regime. (a) Mean $\mathrm{r}\ell_2$ error over the 200-step rollout with and without the embedded-physics component. Without this component, the model reduces to the FNO baseline, and the values coincide with those in Table~\ref{tab:metrics_comparison}. (b) $\mathrm{r}\ell_2$ error over the rollout on the Burgers benchmark with and without the low-pass filter. Both panels use a logarithmic scale.}
    \label{fig:ablation_full}
\end{figure*}

To trace the gains in Section~\ref{sec:results} to the components of the proposed model, we perform ablations in both regimes. In the full-physics regime, the proposed model without its embedded-physics component reduces to the neural-operator component trained alone, which is the FNO baseline in Table~\ref{tab:metrics_comparison}. Figure~\ref{fig:ablation_full}(a) compares the two. Without the embedded-physics component, the $\mathrm{r}\ell_2$ error increases on all five benchmarks, by a factor of 1.7 to 4.1 on four of them and 39 on advection--Allen--Cahn. This large effect is consistent with the governing terms carrying substantial information about the reaction-driven interface dynamics of this benchmark. The ablation thus indicates that the embedded-physics component contributes substantially to the gains in Table~\ref{tab:metrics_comparison}.

Figure~\ref{fig:ablation_full}(b) shows the effect of removing the low-pass filter on Burgers, with the neural-operator component and the rollout training unchanged. Without the filter, the $\mathrm{r}\ell_2$ error rises to about $0.06$ within the first 25 steps, compared with $0.023$ with the filter. It remains about three times higher up to $t\approx100$ and about 1.8 times higher at $t=200$. Spatial differentiation amplifies the FFGS reconstruction error at wavenumber $\kappa$ by a factor proportional to $\kappa$ for first derivatives and $\kappa^2$ for second derivatives. Without the filter, these amplified errors enter the governing terms at every step, which is consistent with the early separation of the two curves. The filter suppresses this high-wavenumber content before the governing terms are evaluated and thus stabilizes the physics evaluation.

\begin{figure*}[!t]
    \centering
    \includegraphics[width=\linewidth]{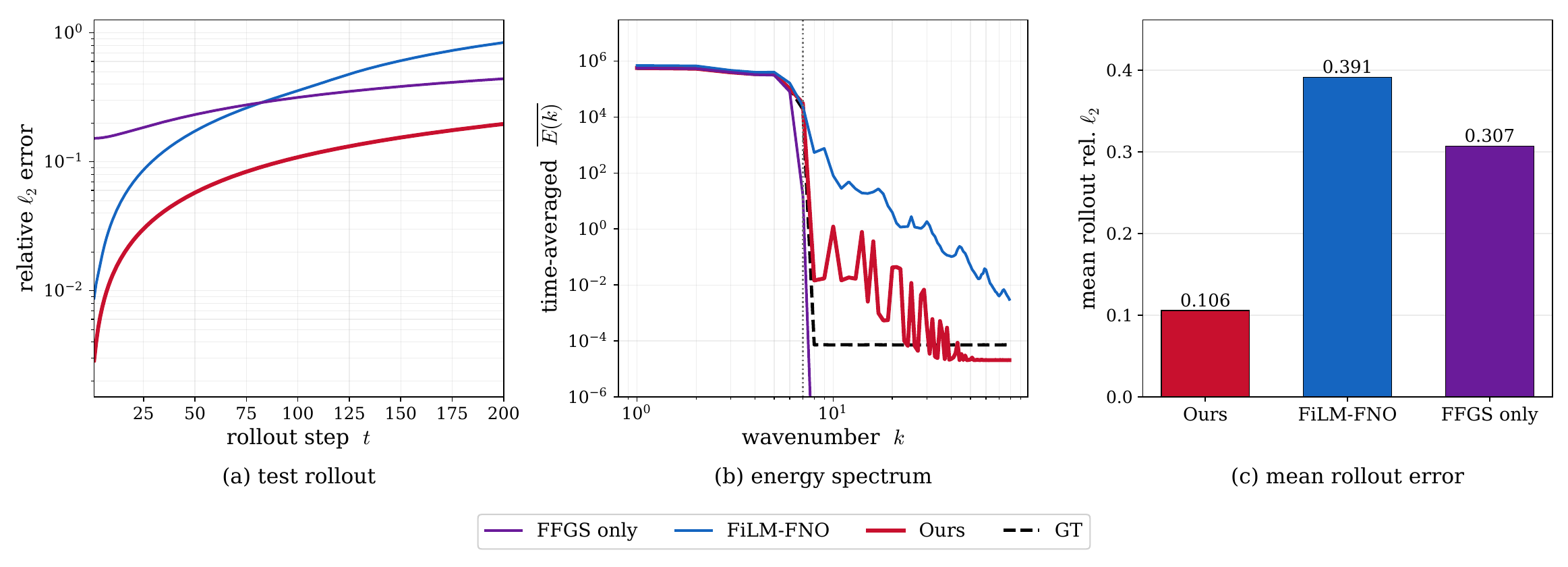}
    \caption{Component ablation in the partial-physics regime. All variants use the same identified coefficients. FFGS-only is the embedded-physics component alone. FiLM-FNO is the neural-operator component trained alone, with the coefficients supplied through FiLM conditioning. (a) $\mathrm{r}\ell_2$ error over the rollout. (b) Time-averaged energy spectra. The dotted line marks the band edge of the initial conditions. (c) Mean $\mathrm{r}\ell_2$ error over the 200-step rollout.}
    \label{fig:partial_ablation}
\end{figure*}

In the partial-physics regime, Fig.~\ref{fig:partial_ablation} compares the proposed model with two variants that use the same identified coefficients. FFGS-only is the embedded-physics component alone. FiLM-FNO is the neural-operator component trained alone, with the coefficients supplied through FiLM conditioning. FFGS-only starts with a large error that grows slowly, whereas FiLM-FNO starts with a small error that grows quickly. The proposed model has the lowest error throughout the rollout, with a mean $\mathrm{r}\ell_2$ of $0.106$ against $0.307$ for FFGS-only and $0.391$ for FiLM-FNO. Supplying the identified coefficients to a neural operator is thus not equivalent to evaluating the corresponding governing terms on the FFGS representation.

Taken together, the ablations indicate that the two components of the proposed model are complementary. The drift of the neural operator alone suggests that the embedded-physics component keeps long rollouts close to the governing dynamics. The offset of the embedded-physics component alone suggests that the accuracy of each step relies mainly on the neural-operator component. Removing the embedded-physics component degrades accuracy on all six benchmarks, which suggests that the gain is a property of the physics-integrated formulation.

\subsection{Limitations and future directions}
\label{sec:limitations}
The capacity of the FFGS encoder limits the method in two respects. First, the number of Gaussians is fixed, with one anisotropic Gaussian per anchor cell and $A=20$ cells per dimension in all experiments. This anchor density bounds the spectral range over which the representation supports accurate physics evaluation. Content beyond this range, such as sharp fronts or persistent high-wavenumber modes, may therefore be reconstructed only approximately, with larger errors in the derivatives than in the field. Adaptive Gaussian representations, such as anchor refinement in regions of high local complexity, could widen this range and reduce the reliance on low-pass filtering. Second, the encoder is trained on a single equation family, so its reconstruction quality, derivative accuracy, and cutoff are tied to that family. Within this family, the test trajectories use new initial conditions drawn from the training distribution, so out-of-distribution initial conditions and coefficients remain untested. Extending the method across multiple PDEs would require a shared representation whose closed-form derivatives, and not only its rendered fields, remain reliable across operators and parameter regimes. A possible direction is to condition the encoder and the neural-operator component on the governing coefficients, nondimensional parameters, or an operator descriptor.

The cutoff wavenumber $\kappa_c$ is selected manually for each benchmark. It is chosen to retain the spectral range in which the FFGS reconstruction and its derivatives are reliable. This choice remains external to the model and involves a trade-off. Removing the filter increases the error on Burgers by up to a factor of about three (Section~\ref{sec:ablations}). At the same time, the energy loss of the FFGS-only variant near the band edge in Fig.~\ref{fig:partial_ablation}(b) suggests that the filter also removes part of the in-band content. A more systematic strategy would estimate the reliable spectral range from diagnostics of the reconstructed field and its derivatives. Since the FFGS reconstruction is explicit, its spectral decay and the errors of the field and of its first and second derivatives can be measured against reference fields (\ref{app:diagnostics}). These diagnostics could select $\kappa_c$ automatically and, in future extensions, adapt it to the field, the governing operator, and the time step.

Finally, the embedded-physics component dominates the inference cost. It accounts for about 12 of the 13.3~ms per step in 2D and 127 of the 129.9~ms in 3D (Section~\ref{sec:cost}), which makes the proposed model 14 and 42 times slower per step than FNO. More efficient rendering and time integration would make the approach more practical for long or large-scale rollouts.

\section{Conclusion}
\label{sec:conclusion}
In this work, we presented a physics-integrated neural operator that incorporates available governing-equation knowledge at the level of the field representation. The representation is built with feed-forward Gaussian splatting (FFGS), which reconstructs each discrete state as a continuous Gaussian field with closed-form spatial derivatives. With these derivatives, an embedded-physics component evaluates the governing terms inside the time stepping of a generic neural operator, without a physics-residual loss or a PDE-specific architecture.

On two- and three-dimensional benchmarks, the proposed model reduces the $\mathrm{r}\ell_2$ error of long-horizon autoregressive rollouts by $1.5\times$--$2.2\times$ relative to the strongest purely data-driven baseline of comparable size and consistently improves spectral fidelity. These gains persist beyond the training horizon and on unseen initial conditions. When the governing form is known but its coefficients are not, the closed-form FFGS derivatives allow the coefficients to be identified from a few observed frames, and the advantage over the baselines persists. Despite the limitations discussed in Section~\ref{sec:limitations}, these results support representation-level physics integration as a practical route for coupling neural operators with fully or partially known PDE structure.

\bibliographystyle{elsarticle-num}
\bibliography{ref}
\appendix

\section{FFGS architecture and implementation}
\label{app:ffgs_impl}

\subsection{Encoder architecture}
\label{app:ffgs_encoder}

The encoder $\mathcal{E}_\varphi$ maps a field with $N^d$ grid points and $C$ channels to $G=A^{d}$ Gaussian primitives, one for each cell of the anchor lattice. All experiments use $A=20$, so that $G=400$ in two dimensions and $G=8000$ in three. The encoder is a four-level U-Net with periodic padding. A $1\times1$ prediction head outputs, for each anchor cell, the center offset, logarithmic scales, orientation, and channel-wise amplitudes of one primitive. The encoder has about 451k trainable parameters for the two-dimensional benchmarks and 1.33M for the three-dimensional benchmark.

\subsection{Field normalization}
\label{app:ffgs_norm}
Each state is normalized channel by channel before encoding. With the time argument omitted, the spatial mean and standard deviation of channel $c$ are
\begin{equation}
m_c=\langle u_c\rangle_\Omega,
\qquad
s_c=\operatorname{std}_\Omega(u_c),
\end{equation}
and the encoder receives
\begin{equation}
\bar{u}_c=\frac{u_c-m_c}{s_c+\varepsilon},
\qquad
\varepsilon=10^{-6}.
\end{equation}
The reconstruction $\hat{\bar{u}}_c$ of the normalized field and its derivatives are mapped back to the physical scale by
\begin{equation}
\hat{u}_c=(s_c+\varepsilon)\,\hat{\bar{u}}_c+m_c,
\qquad
\nabla_{\mathbf{x}}^{n}\hat{u}_c=(s_c+\varepsilon)\,\nabla_{\mathbf{x}}^{n}\hat{\bar{u}}_c,
\quad n=1,2.
\end{equation}
The statistics $m_c$ and $s_c$ are recomputed for every state that enters the encoder, including the intermediate Runge--Kutta states (\ref{app:filter_rk4}).

\subsection{Gaussian parameterization}
\label{app:ffgs_param}
Each primitive is tied to one anchor cell. With anchor spacing $h=1/A$, cell center $\mathbf{x}_{c_i}$, and predicted offset $\mathbf{o}_i$, the center of primitive $i$ is
\begin{equation}
\boldsymbol{\mu}_i=\Big(\mathbf{x}_{c_i}+\frac{h}{2}\tanh(\mathbf{o}_i)\Big)\bmod 1.
\end{equation}
The bounded offset keeps each primitive inside its anchor cell while allowing sub-cell positioning. The covariance follows Eq.~\eqref{eq:ffgs_cov}. The principal scales $\boldsymbol{\sigma}_i$ are obtained by exponentiating the predicted logarithmic scales and clipping them to $[\sigma_{\min},\sigma_{\max}]=[0.008,\,0.25]$. The rotation $R_i$ is parameterized by one angle in two dimensions and by a normalized quaternion in three dimensions. The amplitudes $\boldsymbol{\alpha}_i\in\mathbb{R}^{C}$ are predicted directly.

\subsection{Periodic local rendering}
\label{app:ffgs_render}
All benchmarks are posed on the periodic domain $[0,1)^d$. In Eqs.~\eqref{eq:ffgs_cov} and~\eqref{eq:ffgs_kernel_deriv}, the displacement $\mathbf{x}-\boldsymbol{\mu}_i$ is replaced by its minimum image,
\begin{equation}
\mathbf{r}_i(\mathbf{x})=\mathbf{x}-\boldsymbol{\mu}_i-\operatorname{round}(\mathbf{x}-\boldsymbol{\mu}_i),
\end{equation}
where the rounding acts on each coordinate. The window $\mathcal{S}(q)$ in Eq.~\eqref{eq:ffgs_render} contains the $(2W_{\mathrm{loc}}+1)^d$ anchor cells centered on the cell of $\mathbf{x}_q$, with $W_{\mathrm{loc}}=4$, that is, $9^d$ cells. All primitives in the window enter the sum, without additional distance-based culling. The same window is used in representation training and in the embedded-physics component.

Table~\ref{tab:local_dense_rendering} compares this local rendering with dense summation over all primitives on the two-dimensional benchmarks. The relative discrepancy stays below $3.2\times10^{-5}$ for the field and its derivatives, while the local rendering is 13.9 to 14.6 times faster.

\begin{table}[!t]
\centering
\small
\begin{tabular}{lccccc}
\toprule
& \multicolumn{4}{c}{Relative $\ell_2$ discrepancy} & \\
\cmidrule(lr){2-5}
Benchmark & $\hat{u}$ & $\partial_x\hat{u}$ & $\partial_y\hat{u}$ & $\nabla^2\hat{u}$ & Speedup \\
\midrule
2D Adv              & $1.8\times10^{-6}$ & $5.7\times10^{-6}$ & $1.0\times10^{-5}$ & $2.8\times10^{-5}$ & $13.9\times$ \\
2D Adv--Diff        & $1.9\times10^{-6}$ & $5.3\times10^{-6}$ & $1.2\times10^{-5}$ & $3.2\times10^{-5}$ & $14.3\times$ \\
2D Burgers          & $7.1\times10^{-8}$ & $1.1\times10^{-7}$ & $1.6\times10^{-7}$ & $5.8\times10^{-7}$ & $13.9\times$ \\
2D Adv--Allen--Cahn & $7.5\times10^{-8}$ & $1.1\times10^{-7}$ & $2.3\times10^{-7}$ & $8.7\times10^{-7}$ & $14.6\times$ \\
\bottomrule
\end{tabular}
\caption{Local rendering with $W_{\mathrm{loc}}=4$ compared with dense summation over all primitives. The table reports the relative $\ell_2$ discrepancies of the reconstructed field and its derivatives and the speedup of the local rendering.}
\label{tab:local_dense_rendering}
\end{table}

\section{Representation diagnostics}
\label{app:diagnostics}

We measure the accuracy of the frozen FFGS reconstruction and its closed-form derivatives on each full-physics benchmark, using pseudo-spectral differentiation of the reference fields. For the reconstruction $\hat{u}$ of a field $u$, the relative errors are
\begin{equation}
e_{u}=\frac{\|\hat{u}-u\|_2}{\|u\|_2},
\qquad
e_{\nabla u}=\frac{\|\nabla_{\mathbf{x}}\hat{u}-\nabla_{\mathbf{x}}u\|_2}{\|\nabla_{\mathbf{x}}u\|_2},
\qquad
e_{\nabla^2 u}=\frac{\|\nabla^2_{\mathbf{x}}\hat{u}-\nabla^2_{\mathbf{x}}u\|_2}{\|\nabla^2_{\mathbf{x}}u\|_2}.
\label{eq:derivative_metrics}
\end{equation}
Table~\ref{tab:representation_diagnostics} lists these errors. The field error lies between $1.3\times10^{-3}$ and $5.9\times10^{-3}$ on all benchmarks. Differentiation increases the error, and the Laplacian error exceeds the field error by factors from 2.3 on 3D advection--diffusion to 35 on advection--Allen--Cahn. This growth with derivative order motivates the low-pass filter of Eq.~\eqref{eq:ffgs_filter}, which acts before the derivatives enter the governing terms (\ref{app:filter_rk4}).

\begin{table}[!t]
\centering
\small
\begin{tabular}{lccc}
\toprule
Benchmark & $e_u$ & $e_{\nabla u}$ & $e_{\nabla^2 u}$ \\
\midrule
2D Adv              & $1.41\times10^{-3}$ & $2.57\times10^{-3}$ & $6.51\times10^{-3}$ \\
2D Adv--Diff        & $1.27\times10^{-3}$ & $2.48\times10^{-3}$ & $6.18\times10^{-3}$ \\
2D Burgers          & $1.51\times10^{-3}$ & $5.13\times10^{-3}$ & $2.62\times10^{-2}$ \\
2D Adv--Allen--Cahn & $1.53\times10^{-3}$ & $1.17\times10^{-2}$ & $5.38\times10^{-2}$ \\
3D Adv--Diff        & $5.87\times10^{-3}$ & $8.15\times10^{-3}$ & $1.34\times10^{-2}$ \\
\bottomrule
\end{tabular}
\caption{Relative $\ell_2$ errors of the frozen FFGS reconstruction and its closed-form derivatives, measured against pseudo-spectral references.}
\label{tab:representation_diagnostics}
\end{table}

\section{Filtering and time integration}
\label{app:filter_rk4}

For a grid quantity $f$, the low-pass filter of Eq.~\eqref{eq:ffgs_filter} acts on the discrete Fourier transform of $f$,
\begin{equation}
\Pi_{\kappa_c}f=\mathrm{DFT}^{-1}\!\Big[\,w\big(|\mathbf{q}|\big)\,\mathrm{DFT}[f](\mathbf{q})\Big],
\label{eq:filter_app}
\end{equation}
where $\mathbf{q}$ is the Fourier mode and $w$ is the filter weight. The sharp filter uses
\begin{equation}
w(\kappa)=
\begin{cases}
1, & \kappa\le\kappa_c,\\
0, & \kappa>\kappa_c,
\end{cases}
\end{equation}
and the smooth filter uses
\begin{equation}
w(\kappa)=
\begin{cases}
1, & \kappa\le\kappa_c,\\
\exp\!\big[-(\kappa-\kappa_c)^2/\eta^2\big], & \kappa>\kappa_c,
\end{cases}
\end{equation}
with roll-off width $\eta$.

The filter serves three purposes. It suppresses the high-wavenumber errors of the reconstructed derivatives (\ref{app:diagnostics}), limits the bandwidth seen by the explicit time integrator, and removes out-of-band modes generated by nonlinear products. For Burgers and advection--Allen--Cahn, the assembled right-hand side is therefore filtered again after the nonlinear terms are evaluated. The linear benchmarks omit this second filter, since a linear combination of filtered quantities generates no new modes.

Table~\ref{tab:cutoffs} lists the cutoffs. Each cutoff balances the retention of dynamically relevant modes against the reliability of the FFGS derivatives and the stability of the explicit integration. The initial conditions are band-limited to $k_{\max}=5$ along each axis (Table~\ref{tab:benchmarks}), that is, to radial wavenumbers up to $5\sqrt{2}\approx7$. The linear benchmarks therefore use cutoffs close to this band, whereas the nonlinear benchmarks retain a wider band for the harmonics generated by the dynamics.

\begin{table}[!t]
\centering
\small
\begin{tabular}{lcccc}
\toprule
Benchmark & $N/2$ & $\Delta t$ & $\kappa_c$ & Filter \\
\midrule
2D Adv                         & 80 & 1.0 & 6  & Smooth, $\eta=2$ \\
2D Adv--Diff                   & 80 & 1.0 & 6  & Smooth, $\eta=2$ \\
2D Burgers                     & 80 & 1.0 & 10 & Sharp \\
2D Adv--Allen--Cahn            & 80 & 0.1 & 14 & Sharp \\
3D Adv--Diff                   & 32 & 1.0 & 8  & Smooth, $\eta=2$ \\
\midrule
2D Adv--Diff (partial physics) & 80 & 1.0 & 7  & Sharp \\
\bottomrule
\end{tabular}
\caption{Filter settings of the embedded-physics component. $N/2$ is the largest Fourier index along each axis, and $\Delta t$ is the integration step, which equals the frame spacing.}
\label{tab:cutoffs}
\end{table}

The embedded-physics component advances the state with one classical fourth-order Runge--Kutta (RK4) step per frame interval, without substepping. At each of the four stages, the intermediate state is encoded by the frozen FFGS encoder with recomputed normalization statistics, its derivatives are rendered and filtered, and the right-hand side is assembled. The stages are combined, and the result is filtered once more,
\begin{equation}
\Phi_{\Delta t}^{\mathrm{FFGS}}\big(\mathbf{u}(t);\boldsymbol{\lambda}\big)
=\Pi_{\kappa_c}\Big[\mathbf{u}(t)+\frac{\Delta t}{6}\big(\mathbf{K}_1+2\mathbf{K}_2+2\mathbf{K}_3+\mathbf{K}_4\big)\Big],
\label{eq:filtered_rk4_app}
\end{equation}
where $\mathbf{K}_1,\ldots,\mathbf{K}_4$ are the stage right-hand sides. This final filter restricts the output of the embedded-physics component to the retained band and prevents unresolved content from accumulating over repeated updates. Each update thus involves four encoder evaluations and four renderings. All filters are confined to the embedded-physics component, and the neural-operator component acts on the unfiltered grid state.

\section{Network architectures and training protocol}
\label{app:no_impl}

\subsection{Neural-operator component}
\label{app:no_component}
In the full-physics regime, $\mathcal{N}_\theta$ is the \texttt{ClassicFNO} implementation of \texttt{pdequinox}, the architecture library underlying APEBench~\cite{koehler_apebench_2024}. It has four Fourier blocks with ten retained modes per dimension, six hidden channels, and GELU activations. This configuration has 57,787 trainable parameters for the 2D scalar benchmarks, 57,800 for the two-component Burgers benchmark, and 1,152,187 for the 3D benchmark. The final projection layer is initialized to zero. Before Stage~2, $\mathcal{N}_\theta$ therefore outputs zero, and $\mathcal{G}^{\theta}_{\Delta t}$ coincides with the embedded-physics component.

In the partial-physics regime, $\mathcal{N}_\theta$ is a coefficient-conditioned FNO implemented in Flax. Its number of Fourier blocks, retained modes, and hidden width match the configuration above, and feature-wise linear modulation (FiLM) is applied in every Fourier block. The coefficients are standardized with statistics computed over the training trajectories. For block $\ell$, an affine layer without hidden layers maps the standardized coefficients to a scale $\mathbf{s}^{(\ell)}$ and a bias $\mathbf{b}^{(\ell)}$, each with one entry per hidden channel. The block features $\mathbf{z}^{(\ell)}$ are modulated as
\begin{equation}
\mathbf{z}^{(\ell)}\leftarrow\mathbf{z}^{(\ell)}\odot\big(\mathbf{1}+\mathbf{s}^{(\ell)}\big)+\mathbf{b}^{(\ell)},
\label{eq:film_conditioning_app}
\end{equation}
so that zero modulation recovers the identity map. The final projection layer is again initialized to zero.

\subsection{Baselines}

The baselines follow their APEBench configurations~\cite{koehler_apebench_2024}, and their trainable budgets on the full-physics benchmarks are listed in Table~\ref{tab:metrics_comparison}. In the full-physics regime, the FNO baseline uses the same \texttt{ClassicFNO} configuration as $\mathcal{N}_\theta$ and therefore coincides with the neural-operator component trained alone (Section~\ref{sec:ablations}). All baselines are trained with the Stage~2 settings of Table~\ref{tab:hparams} and the five-step rollout loss of Eq.~\eqref{eq:rollout_loss}, without the embedded-physics component. In the partial-physics regime, the baselines receive the most recent $W=7$ frames as input. The variants evaluated on Burgers and in the partial-physics regime enlarge the trainable budget of each baseline tenfold and are otherwise trained in the same way. On Burgers, the tenfold variants have 579k (FNO), 565k (U-Net), and 612k (ResNet) trainable parameters.

\subsection{Training protocol}

Training has two stages (Fig.~\ref{fig:architecture}). In Stage~1, the FFGS encoder minimizes Eq.~\eqref{eq:ffgs_loss}, and its final checkpoint is frozen. In Stage~2, only $\theta$ is optimized with the five-step rollout loss of Eq.~\eqref{eq:rollout_loss}. During backpropagation, the output of the embedded-physics component is treated as a constant, so the FFGS evaluations and the RK4 integration are not differentiated. This treatment is analogous to the no-gradient unrolling of List et al.~\cite{list_differentiability_2025}, in which gradients are not propagated through the numerical solver. Table~\ref{tab:hparams} lists the optimization settings. Stage~1 decays the learning rate from $1.5\times10^{-3}$ to $10^{-6}$ with a cosine schedule. Stage~2 increases the learning rate linearly to $10^{-3}$ over 2,000 warmup steps and then decays it to zero with a cosine schedule. No validation-based checkpoint selection or early stopping is used.

\begin{table}[!t]
\centering
\small
\begin{tabular}{lcc}
\toprule
 & Stage~1: FFGS encoder & Stage~2: rollout training \\
\midrule
Optimizer & AdamW & Adam \\
Learning-rate schedule & Cosine decay & Linear warmup, cosine decay \\
Peak learning rate & $1.5\times10^{-3}$ & $1.0\times10^{-3}$ \\
Final learning rate & $1.0\times10^{-6}$ & $0$ \\
Training length & 80 epochs & 10,000 steps \\
Batch size & 32 & 20 \\
Rollout length & -- & 5 steps \\
Objective & Eq.~\eqref{eq:ffgs_loss} & Eq.~\eqref{eq:rollout_loss} \\
Checkpoint & Final epoch & Final step \\
\bottomrule
\end{tabular}
\caption{Optimization settings of the two training stages. For the 3D benchmark, Stage~1 uses 40 epochs and batch size 8, and Stage~2 uses a global batch size of 12. For the partial-physics benchmark, Stage~1 uses 100 epochs and batch size 64.}
\label{tab:hparams}
\end{table}

\section{Partial-physics benchmark}
\label{app:peclet}

\subsection{Dataset}
\label{app:partial_data}
The partial-physics benchmark is governed by
\begin{equation}
\partial_t u=-v\,(\partial_x u+\partial_y u)+D\,\nabla^2 u,
\label{eq:advdiff_partial}
\end{equation}
which is the 2D advection--diffusion equation of Table~\ref{tab:benchmarks} with $\mathbf{v}=v\mathbf{1}$ and unknown coefficients $\boldsymbol{\lambda}=(v,D)$. For each trajectory, $\log v$ and $\log D$ are sampled independently and uniformly over the logarithms of $v\in[0.0125,\,0.025]$ and $D\in[10^{-6},\,4\times10^{-6}]$. The P\'eclet number $\mathrm{Pe}=vL/D$, with domain length $L=1$, thus ranges from about $3.1\times10^{3}$ to $2.5\times10^{4}$. The coefficients are constant within a trajectory, and training and test trajectories are sampled independently from the same distributions. The 512 training trajectories span 50 steps. Each of the 30 test trajectories spans 200 steps. Its first $W=7$ frames are observed, and the evaluated rollout covers the remaining 194 steps.

\subsection{Coefficient identification}
\label{app:coef_id}
The coefficients of each trajectory are identified once from its first $W=7$ frames, before Stage~2 training for training trajectories and before the rollout for test trajectories. No later frames are used for identification. With $u_n$ denoting frame $n$, the temporal derivative at the middle frame $n=3$ is estimated by the sixth-order centered difference
\begin{equation}
\widehat{\partial_t u}=\frac{u_6-9u_5+45u_4-45u_2+9u_1-u_0}{60\,\Delta t}.
\label{eq:temporal_derivative_6th}
\end{equation}
The frozen encoder is applied to $u_3$, and the spatial derivatives of its reconstruction are evaluated in closed form without the low-pass filter. The advection coefficient is estimated separately along each axis, with $\boldsymbol{\beta}=(v_x,v_y,D)$ and
\begin{equation}
\mathbf{A}\boldsymbol{\beta}\approx\mathbf{b},
\qquad
\mathbf{A}=\big[-\partial_x\hat{u},\;-\partial_y\hat{u},\;\nabla^2\hat{u}\big],
\qquad
\mathbf{b}=\widehat{\partial_t u},
\label{eq:coefficient_system}
\end{equation}
where the grid points form the rows of $\mathbf{A}$ and $\mathbf{b}$. Each column of $\mathbf{A}$ is normalized by its Euclidean norm, so that $\widetilde{\mathbf{A}}=\mathbf{A}\mathbf{S}^{-1}$ with $\mathbf{S}=\operatorname{diag}\big(\|\mathbf{A}_{:,1}\|_2,\|\mathbf{A}_{:,2}\|_2,\|\mathbf{A}_{:,3}\|_2\big)$. The normalized system is solved by ordinary least squares without regularization,
\begin{equation}
\hat{\boldsymbol{\gamma}}=\operatorname*{arg\,min}_{\boldsymbol{\gamma}}\big\|\widetilde{\mathbf{A}}\boldsymbol{\gamma}-\mathbf{b}\big\|_2^2,
\qquad
\hat{\boldsymbol{\beta}}=\mathbf{S}^{-1}\hat{\boldsymbol{\gamma}}.
\label{eq:coefficient_ls}
\end{equation}
The isotropic velocity is the mean of the two directional estimates, $\hat{v}=(\hat{\beta}_1+\hat{\beta}_2)/2$, and $\hat{D}=\hat{\beta}_3$, which gives $\hat{\boldsymbol{\lambda}}=(\hat{v},\hat{D})$. The identification errors reported in Section~\ref{sec:partial} are the relative errors $|\hat{v}-v|/v$ and $|\hat{D}-D|/D$, averaged over the 30 test trajectories.

\subsection{Use of the identified coefficients}

Both components use the same estimate $\hat{\boldsymbol{\lambda}}$. The embedded-physics component substitutes $(\hat{v},\hat{D})$ into Eq.~\eqref{eq:advdiff_partial}, and $\mathcal{N}_\theta$ receives $\hat{\boldsymbol{\lambda}}$ through FiLM, with standardization statistics computed from the identified coefficients of the training trajectories (\ref{app:no_impl}). The estimates remain fixed during training and prediction, and the ground-truth coefficients are never supplied to either component. Apart from the dataset and the evaluation protocol above, the partial-physics experiments differ from the full-physics ones in three respects. The encoder is trained for 100 epochs with batch size 64 (Table~\ref{tab:hparams}), $\mathcal{N}_\theta$ is the FiLM-conditioned FNO, and the embedded-physics component uses a sharp cutoff with $\kappa_c=7$ (Table~\ref{tab:cutoffs}). The rollout loss, optimizer settings, and checkpoint policy are unchanged.

\section{Supplementary results}
\label{app:supp_results}

Unless stated otherwise, the settings follow Section~\ref{sec:setup}.

\subsection{Rollout error on the remaining full-physics benchmarks}
\label{app:supp_rollout}

\begin{figure*}[!t]
    \centering
    \includegraphics[width=\linewidth]{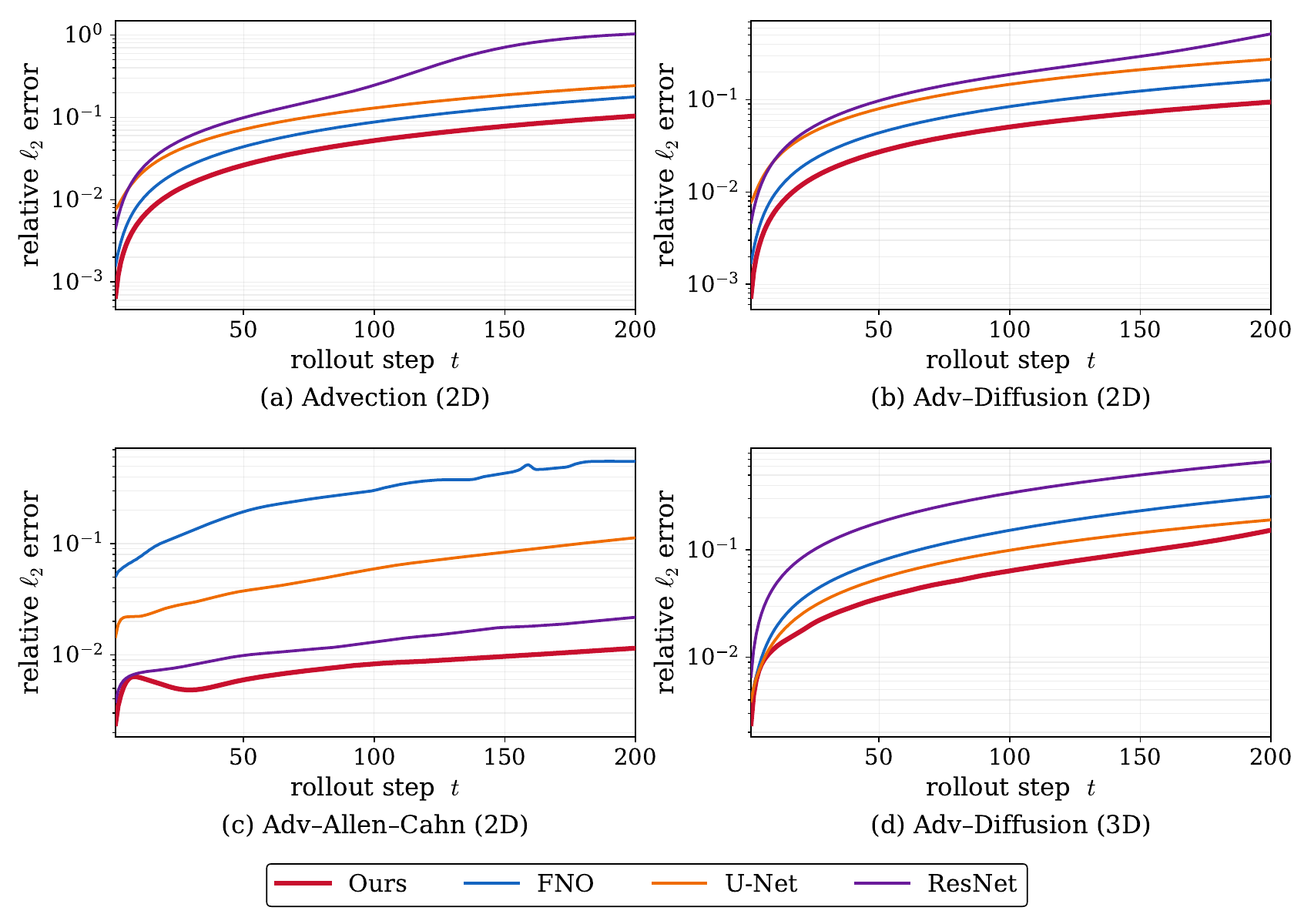}
    \caption{%
    $\mathrm{r}\ell_2$ error over the 200-step rollout on the remaining full-physics benchmarks, averaged over the 30 held-out trajectories and shown on a logarithmic scale. (a) 2D advection, (b) 2D advection--diffusion, (c) 2D advection--Allen--Cahn, and (d) 3D advection--diffusion. All models are trained on the first 50 steps.}
    \label{fig:additional_rollout_curves}
\end{figure*}

Figure~\ref{fig:additional_rollout_curves} extends Fig.~\ref{fig:rollout} to the other four full-physics benchmarks. On 2D advection, 2D advection--diffusion, and 3D advection--diffusion, the proposed model has the lowest $\mathrm{r}\ell_2$ throughout the rollout. On the two 2D linear benchmarks, its error stays 1.4 to 1.8 times below that of FNO, the strongest baseline, from $t=50$ to $t=200$. On 3D advection--diffusion, it stays below U-Net, the strongest baseline there, although the margin narrows from about 1.5 at $t=100$ to 1.2 at $t=200$. On advection--Allen--Cahn, the proposed model and ResNet are several times more accurate than U-Net and FNO. The two are nearly equal around $t\approx10$, after which the error of the proposed model stays 1.6 to 1.9 times below that of ResNet. The advantage of the proposed model is therefore not specific to Burgers, consistent with Table~\ref{tab:metrics_comparison}.

\begin{figure*}[!t]
    \centering
    \includegraphics[width=\linewidth]{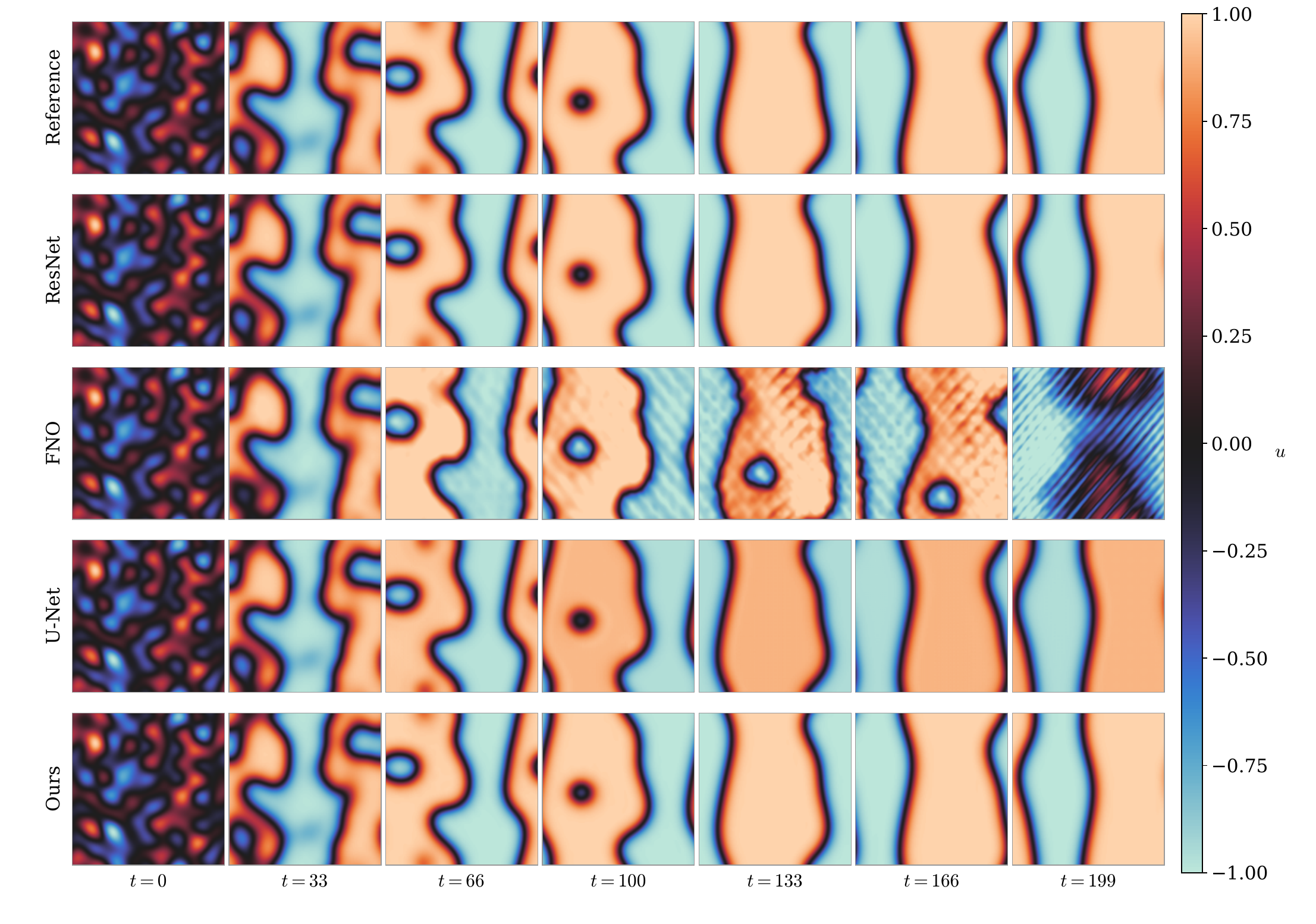}
    \caption{%
    Representative rollout on 2D advection--Allen--Cahn. Columns show the initial condition and the predictions at $t=33$, $66$, $100$, $133$, $166$, and $199$. Rows show the reference solution, ResNet, FNO, U-Net, and the proposed model. The order parameter $u$ uses a common color scale across all methods and lead times.}
    \label{fig:snapshot_ac}
\end{figure*}

Figure~\ref{fig:snapshot_ac} shows the order parameter on advection--Allen--Cahn. All methods reproduce the initial coarsening of the phase-field structures. FNO develops oscillatory artifacts from $t\approx66$, retains a droplet that has disappeared from the reference, and departs qualitatively from the reference by $t=199$. U-Net reproduces the interface positions but underestimates the magnitude of the order parameter in the bulk phases. ResNet and the proposed model remain visually indistinguishable from the reference. Their difference appears in the $\mathrm{r}\ell_2$, which is 1.6 to 1.9 times lower for the proposed model in Fig.~\ref{fig:additional_rollout_curves}(c).

\begin{figure*}[!t]
    \centering
    \includegraphics[width=\linewidth]{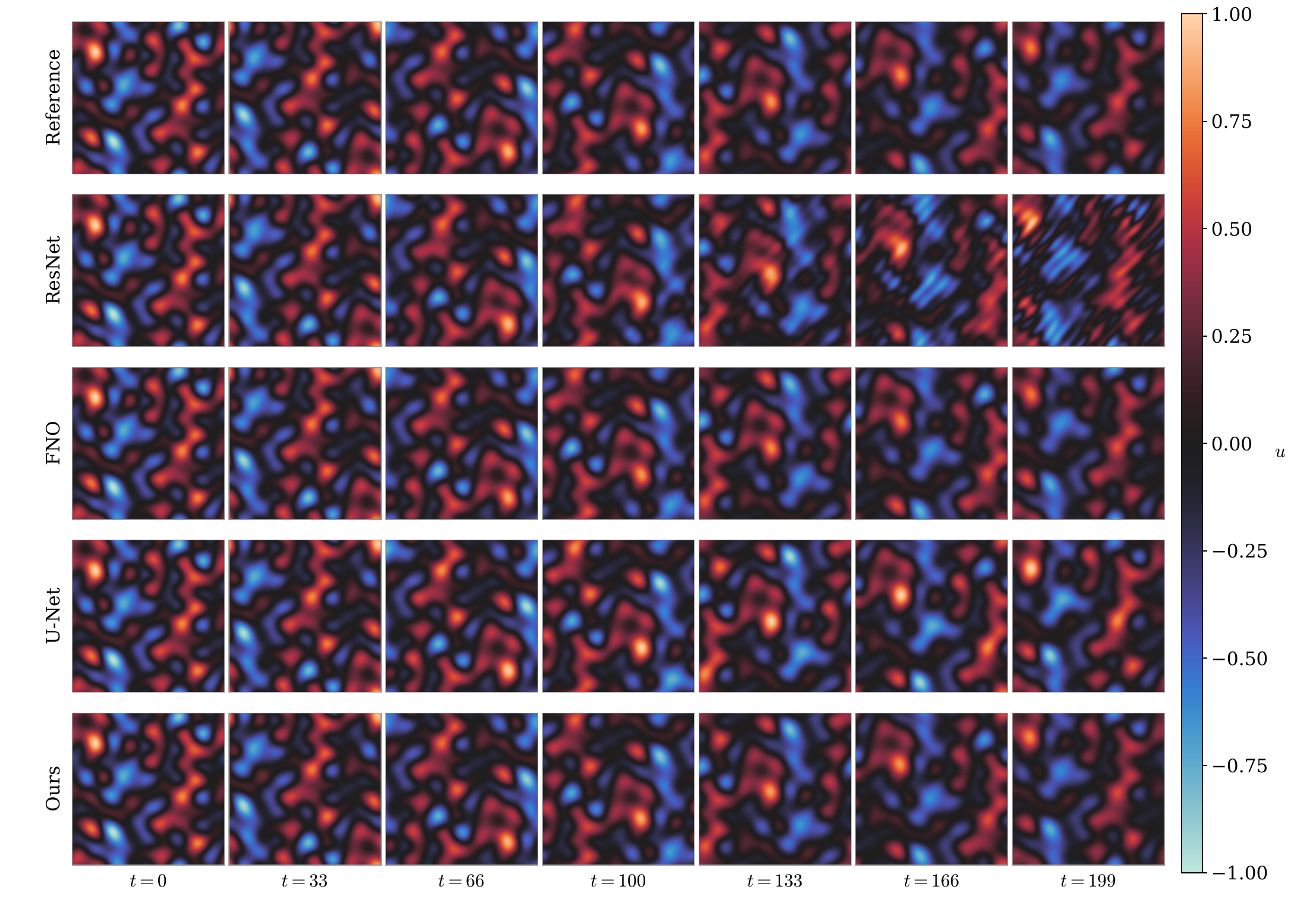}
    \caption{%
    Representative rollout on 2D advection--diffusion. Columns show the initial condition and the predictions at $t=33$, $66$, $100$, $133$, $166$, and $199$. Rows show the reference solution, ResNet, FNO, U-Net, and the proposed model. The scalar field $u$ uses a common color scale across all methods and lead times.}
    \label{fig:snapshot_ad}
\end{figure*}

Figure~\ref{fig:snapshot_ad} shows the scalar field on 2D advection--diffusion. The methods are visually similar up to $t\approx100$. ResNet then develops oscillatory stripes that corrupt the field by $t=199$, consistent with its spectral peak near $\kappa\approx13$ in Fig.~\ref{fig:additional_spectra}(c). U-Net changes the morphology and intensity of individual structures, and FNO accumulates local discrepancies. The proposed model agrees most closely with the reference over the full rollout, including the locations, signs, and characteristic scales of the dominant structures.

\begin{figure*}[!t]
    \centering
    \includegraphics[width=\linewidth]{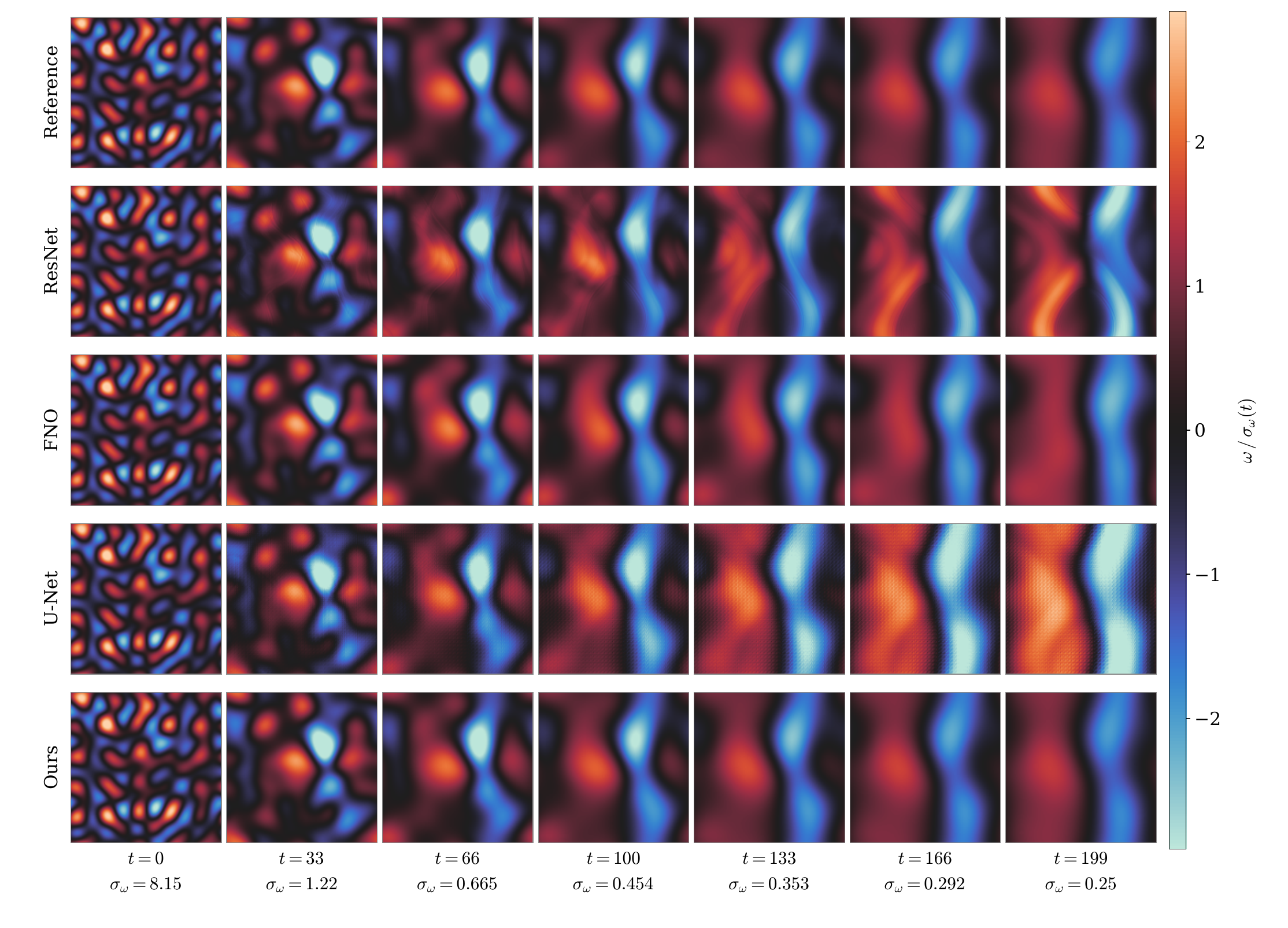}
    \caption{%
    Representative rollout on 2D Burgers. Columns show the initial condition and the predictions at $t=33$, $66$, $100$, $133$, $166$, and $199$. Rows show the reference solution, ResNet, FNO, U-Net, and the proposed model. The displayed quantity is the vorticity normalized by the instantaneous reference standard deviation, $\omega/\sigma_{\omega}(t)$, and the values of $\sigma_{\omega}(t)$ are given below each column.}
    \label{fig:snapshot_burgers}
\end{figure*}

Figure~\ref{fig:snapshot_burgers} shows the normalized vorticity on Burgers. As the flow evolves, the initial small-scale structures merge into smoother and more elongated vortical regions. At later times, ResNet develops spurious filaments, and U-Net overestimates the amplitude and extent of the structures and develops a grid-scale checkerboard texture. FNO remains visually close to the reference. In Fig.~\ref{fig:burgers2d_spectrum_all}(b), it matches the reference energy within a few percent for $2\le\kappa\le7$, the scales that dominate the vorticity. Its overestimation of the energy at $\kappa=1$ by 13 percent is consistent with its larger $\mathrm{r}\ell_2$ in Fig.~\ref{fig:rollout}. The proposed model tracks the location, orientation, and width of the dominant structures most faithfully through $t=199$, in agreement with its lowest $\mathrm{r}\ell_2$ in Fig.~\ref{fig:rollout}.

Across the three examples, several baselines show visible structural failures, such as oscillatory artifacts, spurious filaments, or biased amplitudes, whereas the proposed model preserves the reference structures throughout. These failures appear mostly at lead times well beyond the 50-step training horizon.

\subsection{Spectra on the remaining full-physics benchmarks}
\label{app:supp_spectra}

\begin{figure*}[!t]
    \centering
    \includegraphics[width=\linewidth]{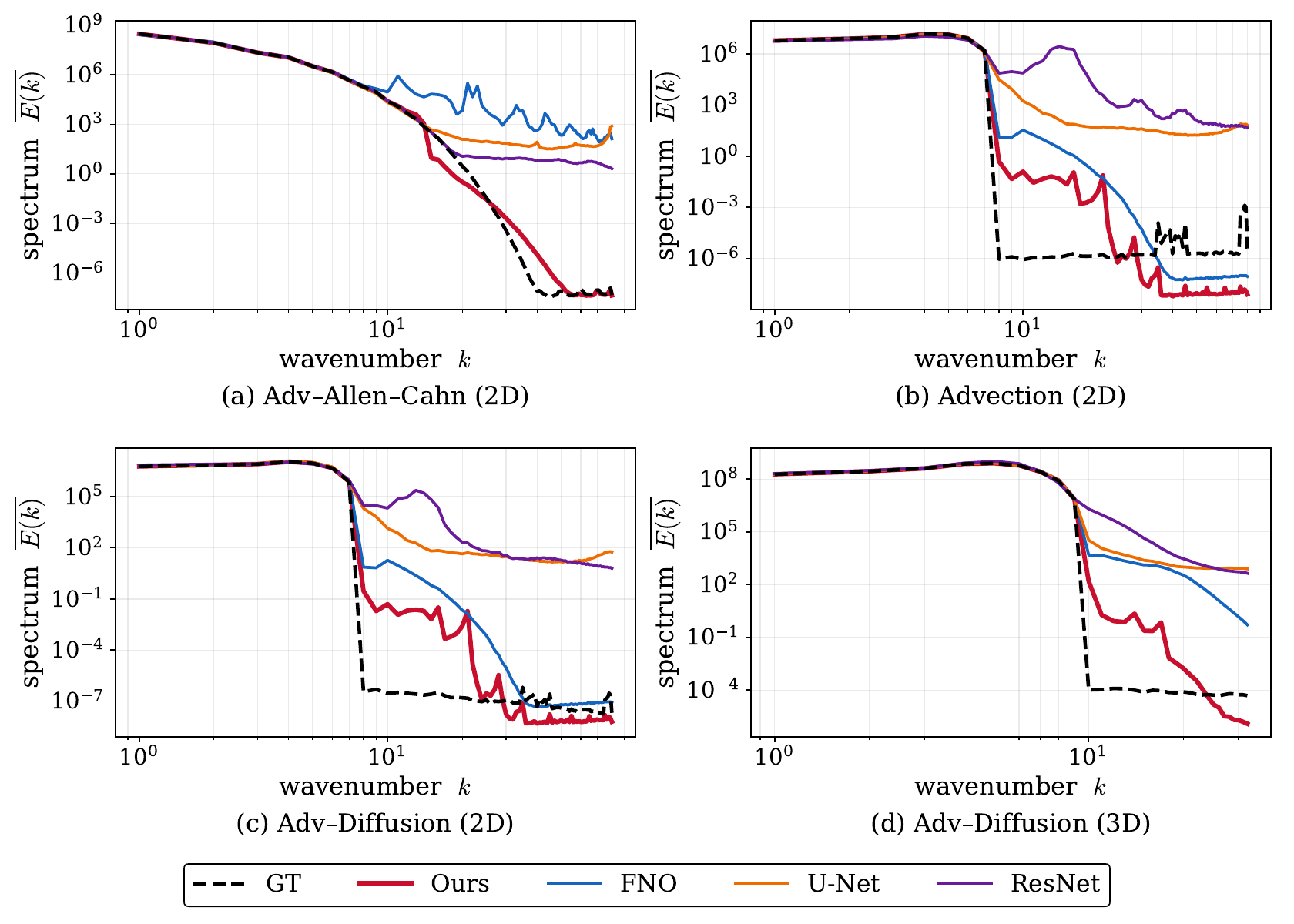}
    \caption{%
    Time-averaged radial energy spectra on the remaining full-physics benchmarks, averaged over the 30 held-out trajectories and 200 rollout steps. (a) 2D advection--Allen--Cahn, (b) 2D advection, (c) 2D advection--diffusion, and (d) 3D advection--diffusion. The $\kappa=0$ mode is retained only for advection--Allen--Cahn (Section~\ref{sec:setup:metrics}).}
    \label{fig:additional_spectra}
\end{figure*}

Figure~\ref{fig:additional_spectra} extends Fig.~\ref{fig:burgers2d_spectrum_all}(a) to the other four full-physics benchmarks. As on Burgers, the proposed model has the lowest spectral error on all four, 3.3 to 8.2 times lower than that of the strongest baseline of each benchmark (Table~\ref{tab:metrics_comparison}). On the three linear benchmarks, its spectrum drops sharply at the band edge of the initial conditions, stays below those of all baselines beyond it, and reaches the reference floor at the highest wavenumbers. On advection--Allen--Cahn, the baselines level off between about $10^{1}$ and $10^{5}$ at high wavenumbers, whereas the proposed model follows the decay of the reference down to its floor near $\kappa\approx50$. Its spectrum falls below the reference just above the cutoff $\kappa_c=14$, similar to the drop just above $\kappa_c=10$ on Burgers in Fig.~\ref{fig:burgers2d_spectrum_all}(b).

\subsection{Model-capacity control on Burgers}
\label{app:supp_capacity}

\begin{figure*}[!t]
    \centering
    \includegraphics[width=\linewidth]{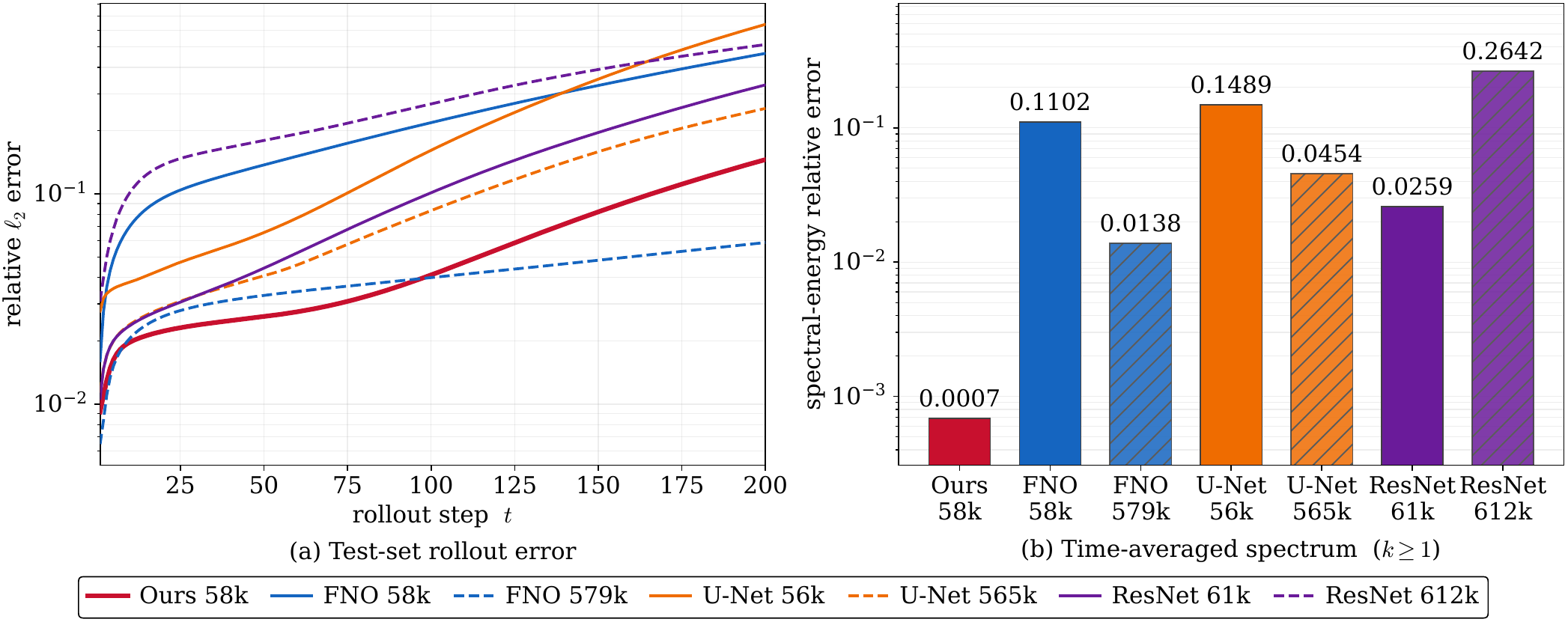}
    \caption{%
    Model-capacity control on Burgers. (a) $\mathrm{r}\ell_2$ error over the rollout for the proposed model and the baselines at matched (solid) and tenfold (dashed) trainable budgets, with parameter counts in the legend. (b) Spectral error, evaluated over $\kappa\ge1$ (Section~\ref{sec:setup:metrics}).}
    \label{fig:capacity_control}
\end{figure*}

Figure~\ref{fig:capacity_control} tests whether model capacity alone explains the gains of the proposed model on Burgers. The trainable budgets of FNO, U-Net, and ResNet are increased tenfold, while the proposed model keeps its configuration with 57.8k trainable parameters. In Fig.~\ref{fig:capacity_control}(a), the tenfold budget improves FNO and U-Net but degrades ResNet. FNO $10\times$ becomes the most accurate model in the second half of the rollout. Its error falls below that of the proposed model after $t\approx100$ and reaches about 0.06 at $t=200$, compared with 0.14 for the proposed model. Between $t\approx10$ and $t\approx100$, the proposed model remains more accurate than FNO $10\times$ with a tenth of its trainable parameters. In Fig.~\ref{fig:capacity_control}(b), the proposed model has the lowest spectral error of all models, more than 20 times below that of FNO $10\times$. The tenfold budget reduces the spectral error of FNO and U-Net but increases that of ResNet. A tenfold FNO thus surpasses the proposed model in late-time $\mathrm{r}\ell_2$ on Burgers, but not in spectral fidelity.

\end{document}